\RequirePackage[bookmarksnumbered,unicode]{hyperref}

\documentclass[sigplan,twocolumn]{acmart}
\usepackage[english]{babel}
\usepackage{blindtext}
\usepackage{color}
\usepackage{multirow}
\usepackage{url}
\usepackage{graphicx}
\usepackage{algorithm}
\usepackage[noend]{algpseudocode}
\usepackage{booktabs}

\usepackage{enumitem} % for concise list env
\usepackage{latexsym}
\usepackage{amsmath}
\usepackage{amsfonts}
\usepackage{tabularx}
\usepackage{balance}
\usepackage{caption}
\usepackage{array}
\setitemize{noitemsep,topsep=1pt,parsep=1pt,partopsep=1pt}
\setenumerate{noitemsep,topsep=1pt,parsep=1pt,partopsep=1pt}

\setlist{nolistsep}

\newcommand{\parabf}[1]{\medskip\noindent\textbf{#1}}

\newcommand{\paraf}[1]{\noindent\textbf{#1}}
\newcommand{\cut}[1]{}

\newcommand{\sysname}{MegaScale-MoE\xspace}

\newcommand{\revision}[1]{\textcolor{black}{#1}}

\setcopyright{acmlicensed}
\acmYear{2026}\copyrightyear{2026}
\acmConference[EUROSYS '26]{European Conference on Computer Systems}{April 27--30, 2026}{Edinburgh, Scotland Uk}
\acmBooktitle{European Conference on Computer Systems (EUROSYS '26), April 27--30, 2026, Edinburgh, Scotland Uk}
\acmDOI{10.1145/3767295.3769325}
\acmISBN{979-8-4007-2212-7/26/04}

\begin{document}
% \title{Communication-Efficient Training of Mixture-of-Experts Models at Scale}

\title{\sysname: Large-Scale Communication-Efficient Training of Mixture-of-Experts Models in Production}

%\titlenote{Produces the permission block, and copyright information}
%\subtitle{Extended Abstract}

% \author{Paper \#XX, \pageref{sec:conclusion} pages}

\newcommand*{\affmark}[1][*]{\textsuperscript{#1}}
\newcommand*{\affaddr}[1]{#1}

\author{
    Chao Jin*\affmark[\S], Ziheng Jiang*\affmark[$\dag$], Zhihao Bai\affmark[$\dag$], Zheng Zhong\affmark[$\dag$], Juncai Liu\affmark[$\dag$], Xiang Li\affmark[$\dag$], Ningxin Zheng\affmark[$\dag$], Xi Wang\affmark[$\dag$], Cong Xie\affmark[$\dag$], Qi Huang\affmark[$\dag$], Wen Heng\affmark[$\dag$], Yiyuan Ma\affmark[$\dag$], Wenlei Bao\affmark[$\dag$], Size Zheng\affmark[$\dag$], Yanghua Peng\affmark[$\dag$], Haibin Lin\affmark[$\dag$], Xuanzhe Liu\affmark[\S], Xin Jin\affmark[\S], Xin Liu\affmark[$\dag$] \\
\affaddr{\affmark[\S]{\textit{School of Computer Science, Peking University}}}~~~~\affaddr{\affmark[$\dag$]{\textit{ByteDance Seed}}}
}

\begin{CCSXML}
<ccs2012>
   <concept>
       <concept_id>10010520.10010521.10010537.10003100</concept_id>
       <concept_desc>Computer systems organization~Cloud computing</concept_desc>
       <concept_significance>500</concept_significance>
       </concept>
   <concept>
       <concept_id>10010147.10010257</concept_id>
       <concept_desc>Computing methodologies~Machine learning</concept_desc>
       <concept_significance>500</concept_significance>
       </concept>
   <concept>
       <concept_id>10003033.10003106.10003110</concept_id>
       <concept_desc>Networks~Data center networks</concept_desc>
       <concept_significance>500</concept_significance>
       </concept>
 </ccs2012>
\end{CCSXML}

\ccsdesc[500]{Computer systems organization~Cloud computing}
\ccsdesc[500]{Computing methodologies~Machine learning}
\ccsdesc[500]{Networks~Data center networks}

\keywords{Mixture-of-experts, distributed training, computation-communication overlap}

\renewcommand{\shortauthors}{C. Jin, et al}
\renewcommand{\shorttitle}{\sysname: Large-Scale Communication-Efficient Training...}
\begin{sloppypar}
\begin{abstract}

We present \sysname, a production system tailored for the efficient training of large-scale 
mixture-of-experts (MoE) models. MoE emerges as a promising architecture to scale large language 
models (LLMs) to unprecedented sizes, thereby enhancing model performance. However, existing MoE 
training systems experience a degradation in training efficiency, exacerbated by the escalating scale 
of MoE models and the continuous evolution of hardware. 
    
Recognizing the pivotal role of efficient communication in enhancing MoE training, \sysname customizes 
communication-efficient parallelism strategies for attention and FFNs in each MoE layer and adopts a 
holistic approach to overlap communication with computation at both inter- and intra-operator levels. 
Additionally, \sysname applies communication compression with adjusted communication patterns to lower 
precision, further improving training efficiency. When training a 352B MoE model on 1,440 NVIDIA Hopper 
GPUs, \sysname achieves a training throughput of 1.41M tokens/s, improving the efficiency by 
1.88$\times$ compared to Megatron-LM. We share our operational experience in accelerating MoE training 
and hope that by offering our insights in system design, this work will motivate future research in 
MoE systems.
    
\end{abstract}

\maketitle
{\let\thefootnote\relax\footnote{{$^*$Equal contribution.}}}

\vspace{-6pt}
{\par\smallskip\small\noindent{\bfseries ACM Reference Format:}\par\nobreak
  \noindent Chao Jin, Ziheng Jiang, Zhihao Bai, Zheng Zhong, Juncai Liu, Xiang Li, Ningxin Zheng, Xi Wang, Cong Xie, Qi Huang, Wen Heng, Yiyuan Ma, Wenlei Bao, Size Zheng, Yanghua Peng, Haibin Lin, Xuanzhe Liu, Xin Jin, Xin Liu. 2025. MegaScale-MoE: Large-Scale Communication-Efficient Training of Mixture-of-Experts Models in Production. 
  In \textit{EuroSys ’26, April
27–30, 2026, Edinburgh, UK.} In  ACM, New York, NY, USA, 17 pages. 
\url{https://doi.org/10.1145/3767295.3769325}
}

\section{Introduction}
\label{sec:introduction}

%Large Language Models (LLMs)~\cite{chowdhery2023palm, touvron2023llama, jiang2024mixtral} have emerged as a cornerstone of modern artificial intelligence research, showcasing unparalleled capabilities in generating human-like text, understanding complex queries, and facilitating groundbreaking advancements across numerous domains~\cite{zhang2023prompting, zhang2024benchmarking, siriwardhana2023improving}. The significance of LLMs is underscored by their increasing role in a wide array of applications, from enhancing natural language processing tasks~\cite{adiwardana2020towards, see2017get, yang2016review} to driving innovation in generative AI technologies~\cite{siriwardhana2023improving, github_copilot, cursor}.

As the size of Large Language Models (LLMs)~\cite{chowdhery2023palm, touvron2023llama, jiang2024mixtral} grow, so does the scale of their training regimes. The escalation in training scale has made efficiency improvements not just desirable but crucial~\cite{jiang2024megascale}. As a company building AI products for billions of users, we remain committed to training LLMs with hundreds of billions of parameters on thousands of GPUs. Consequently, even marginal gains in training efficiency can significantly reduce computational resource consumption and training time, directly influencing the feasibility and sustainability of developing state-of-the-art LLMs.

Within the landscape of LLM architectures, Mixture-of-Experts (MoE) models stand out for their sparse activation~\cite{chowdhery2023palm, jiang2024mixtral, fedus2022switch, shazeer2017outrageously}, which dynamically routes input tokens to a selected set of specialized network components, known as \emph{experts}, rather than to all parameters. This design leads to sub-linear scaling of FLOPs required as the model size increases, thereby significantly reducing the computational cost. 
Recent industrial advancements~\cite{google_GLaM, rajbhandari2022deepspeed, dbrx, grok, liu2024deepseekv3} have demonstrated the potential of MoE models, achieving an order-of-magnitude reduction in training cost compared to dense models with equivalent model quality.

Despite the lower training costs of MoE models, we observe a critical performance bottleneck during training from a systems perspective---communication.
For instance, when training an internal model on NVIDIA Hopper GPUs, communication accounts for 43.6\% of the total time during the forward pass and 32\% over the entire training process. 
Two primary factors contribute to this bottleneck. First, MoE models inherently introduce more communication overhead. Compared to dense model training, MoE model training requires distribution across more GPUs for model parallelism due to its larger parameter size. Second, enabling sparse computation requires two extra all-to-all communications in both the forward and backward passes to dispatch and aggregate tokens, respectively, which hinders ongoing computation.

\begin{figure}[t!]
    \centering
    \includegraphics[width=0.75\linewidth]{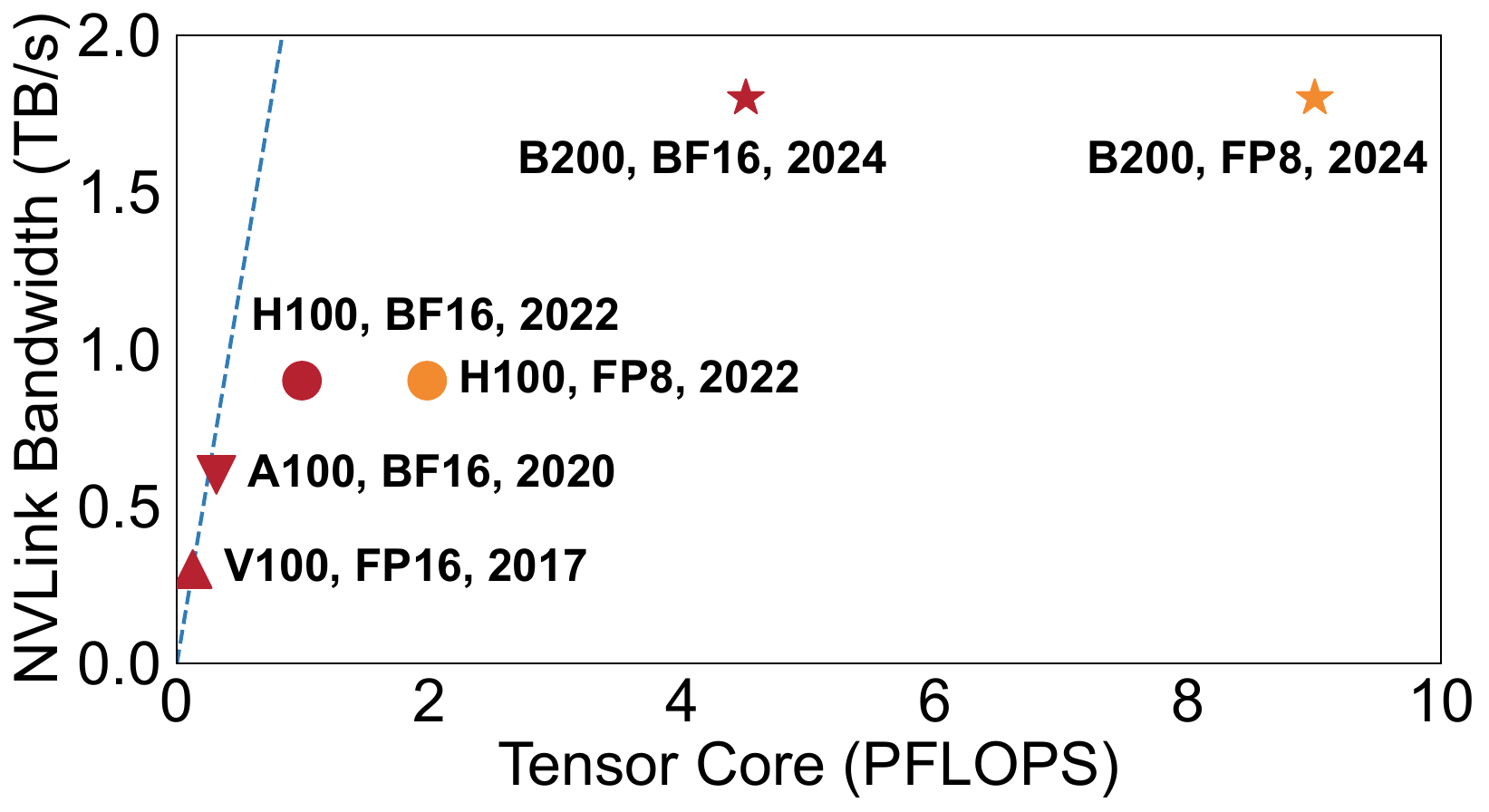}
    \vspace{-0.1in}
    \caption{Evolution of NVIDIA GPUs.}
    \vspace{-0.2in}
    \label{fig:gpu_evolution}
\end{figure}

Moreover, as hardware advances, the imbalance between computation and communication becomes increasingly pronounced, with communication overhead growing more dominant. Alongside improvements in model architectures, hardware capabilities have evolved rapidly, with GPUs achieving significantly higher processing speeds (Figure~\ref{fig:gpu_evolution}). Concurrently, reductions in training precision have been adopted to enhance efficient and cost-effective training~\cite{peng2023fp8, liu2024deepseekv3}. These trends lead to a scenario where the raw computation time decreases, making the relative impact of communication overhead a more critical bottleneck. For instance, simply extending existing tensor parallelism to multi-node setups has been observed to push communication overhead beyond 50\% in certain cases. As a result, optimizing communication is essential for sustaining and improving the scalability of MoE model training, particularly in distributed environments where frequent data synchronization across multiple GPUs is required. 

In this paper, we present the design, implementation, and operational experience of \sysname, a production system optimized for efficient large-scale MoE training. By meticulously addressing the communication bottleneck, \sysname strives to push the boundaries of MoE training, achieving significant improvements in performance and efficiency. 
\revision{Based on the insight that the key architectural distinctions between MoE and dense models are intra-layer, which is the primary source of the communication overhead, \sysname confines each MoE layer to within a single node, utilizing high-bandwidth NVLink. Our analysis (\S\ref{sec:design:parallelism}) and evaluation (\S\ref{sec:evaluation}) show that despite the cross-node expert parallelism common in existing systems~\cite{liu2024deepseekv3, hwang2023tutel}, our approach effectively scales MoE training to models of several hundred billion parameters on thousands of GPUs.}

Specifically, \sysname addresses the communication problem in MoE training from three key aspects. First, \sysname reduces the communication volume by customizing parallelism strategies for the attention and FFN modules in each MoE layer. We compare the parallelism strategies in existing LLM training frameworks, comprehensively considering their impact on large-scale training, including the communication volume and whether communication can be effectively overlapped (i.e., whether it lies on the critical path).
% and their effect on training stability. 
Based on this analysis, we select the optimal combination of parallelism strategies for MoE training. 

Second, \sysname fully overlaps communication with computation at the operator level. \sysname partitions the forward and backward passes of each MoE layer into distinct computation and communication operators. For inter-operator overlap, \sysname employs a holistic 
scheduling strategy that carefully reorders communication and computation operators during both forward and backward propagation, hiding communication within independent computations. This approach also optimizes GPU memory usage. \sysname utilizes selective activation rematerialization, retaining only a subset of activations in GPU memory during the forward pass, and recomputing or re-communicating to obtain the required activations during the backward pass. With this holistic scheduling, \sysname effectively hides the rematerialization overhead, achieving comparable performance while storing only half of the activations. 

To overlap communication on the critical paths, \sysname employs a fine-grained approach that splits communication into tiles and aligns with the GPU compute pattern, fusing these tile-level communications into the compute kernels. For MoE models with token dispatch, \sysname fuses an efficient local scatter operation into the kernel and reorganizes the computation tasks along the scattered dimension to mitigate communication bottlenecks from multiple data sources. This fine-grained overlap occurs within each node, leveraging the high-bandwidth connectivity between GPUs. 

Third, \sysname leverages communication compression to further enhance MoE training efficiency. Specifically, for widely-used BF16 mixed-precision training, \sysname reduces the inter-node parameter synchronization precision from FP32 to BF16, halving the associated overhead. In FP8 training, \sysname replaces BF16 reduce-scatter with FP8 communication, incorporating tailored quantization strategies and FP32 reduction to decrease communication volume while preserving convergence stability.

\sysname is deployed in our datacenters to train MoE models for our products. Compared to the state-of-the-art open-source LLM training framework, Megatron-LM~\cite{shoeybi2019megatron}, \sysname achieves up to 1.88$\times$ higher MFU (Model FLOPs Utilization) when training a 352B MoE model on 1,440 NVIDIA Hopper GPUs. With comprehensive communication  optimizations, \sysname powers large-scale training in our production, efficiently scaling to trillions of parameters and thousands of GPUs while saving millions of GPU hours.
\section{Background}
\label{sec:background}

\begin{figure}[t!]
    \centering
    \includegraphics[width=0.65\linewidth]{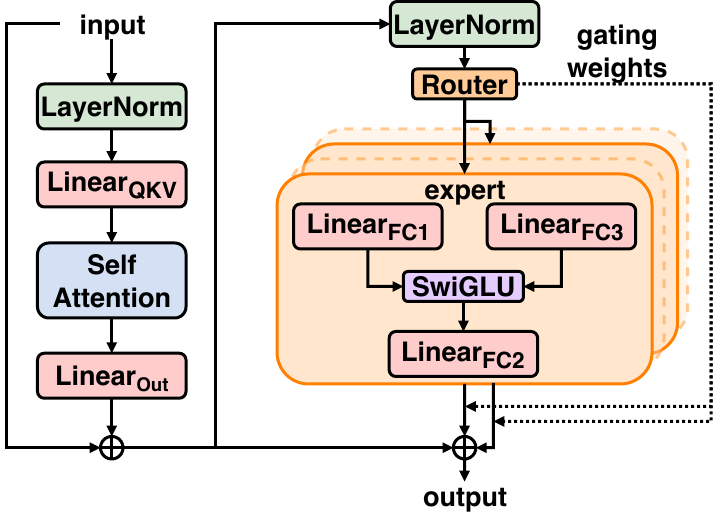}
    \vspace{-0.15in}
    \caption{Mixture-of-Experts (MoE) layer.}
    \vspace{-0.1in}
    \label{fig:moe_intro}
\end{figure}

\subsection{Mixture-of-Experts for Transformer}

The Mixture of Experts (MoE) mechanism is an advanced approach designed to boost the performance of Transformer~\cite{vaswani2017attention} models, which are increasingly pivotal in the realm of LLMs~\cite{jiang2024mixtral, chowdhery2023palm, dbrx, liu2024deepseekv3}. It extends the Transformer architecture by integrating multiple expert networks within the feed-forward network (FFN) component. As illustrated in Figure~\ref{fig:moe_intro}, MoE models dynamically route input tokens to the most relevant experts based on their characteristics. This routing is managed by a trainable gating mechanism that selects the best-suited experts for each token. This architectural innovation enables MoE models to scale in capacity without a proportional increase in inference costs, as only a subset of experts is activated for each input.
% It offers a more flexible and efficient way to improve model performance beyond simply increasing the size of the network.

\subsection{Large-scale LLM Training}

Training large language models at scale on tens of thousands of GPUs is a complex system engineering challenge that requires multiple systems techniques. To distribute the training workload, a combination of parallelism strategies such as data, tensor, and pipeline parallelism is necessary~\cite{shoeybi2019megatron, rasley2020deepspeed, jiang2024megascale}, as each approach has limitations that prevent relying on a single method for effective scaling. 

\parabf{Data parallelism} uniformly distributes the training data across all devices, with each device replicating the model parameters and optimizer states. To synchronize the parameters after each training iteration, data parallelism performs an all-reduce communication operation. Zero Redundancy Optimizer (ZeRO)~\cite{rajbhandari2020zero} improves over data parallelism by distributing model states across all participating devices. ZeRO unfolds across three progressive stages, each designed to increasingly conserve memory, though this comes with the trade-off of elevated communication. 

\begin{figure}[t!]
    \centering
    \includegraphics[width=\linewidth]{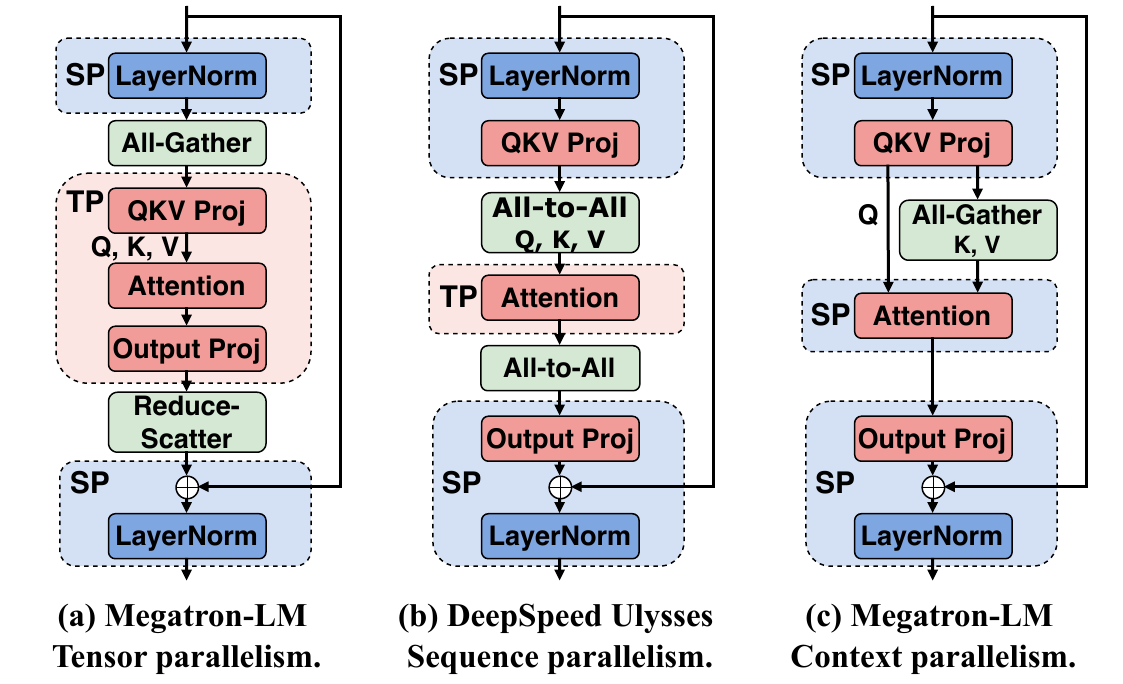}
    \vspace{-0.3in}
    \caption{Different parallelism strategies for self-attention. "TP" denotes partitioning along the dimension of hidden size, while "SP" denotes partitioning along the dimension of sequence length.}
    \vspace{-0.1in}
    \label{fig:background:attention_parallelism}
\end{figure}

\parabf{Tensor parallelism} distributes compute-intensive tensor operations over multiple devices, enabling parallel computation and significantly accelerating the training process. The specific partitioning strategy and the dependencies among operators within the model dictate that tensor parallelism may necessitate gathering split inputs (all-gather) or merging outputs (reduce-scatter). In LLM training, operators like LayerNorm and Dropout, though less compute-intensive, require substantial activation memory. To tackle this problem, a variant of tensor parallelism known as \textbf{sequence parallelism}~\cite{korthikanti2023reducing} is proposed, which partitions these operators along the dimension of sequence length. For long-context training, several works~\cite{shoeybi2019megatron, jacobs2023deepspeed, context_parallelism} apply sequence parallelism or tensor parallelism to different operators in self-attention. Figure~\ref{fig:background:attention_parallelism} illustrates the mainstream parallelism strategies for attention, namely tensor, sequence, and context parallelism (TP, SP, and CP), which we analyze in \S\ref{sec:design:parallelism:sp_attention}.

\parabf{Pipeline parallelism} enhances efficiency by dividing model layers into stages that are processed on different devices, enabling pipelined execution. Each batch is split into several micro-batches for this purpose. To minimize pipeline bubbles, various scheduling strategies have been developed, e.g., GPipe~\cite{huang2019gpipe}, PipeDream 1F1B~\cite{narayanan2019pipedream} and Interleaved 1F1B~\cite{narayanan2021efficient}, etc. Megatron-LM adopts Interleaved 1F1B pipeline scheduling, further dividing each stage on one device into multiple virtual stages to reduce the pipeline bubble rate.

\parabf{Expert parallelism} is tailored for training MoE models by distributing experts across multiple devices, alleviating memory pressure and enabling parallel processing. To efficiently assign tokens to the appropriate experts and retrieve their outputs, all-to-all communication is typically employed.

\iffalse

\begin{table*}[t!]
    \centering
    \resizebox{\linewidth}{!} {
        \begin{tabular}{c|c|c|c|c|c|c}
            \hline
            \hline
            \multirow{2}{*}{\begin{tabular}[c]{@{}c@{}}Parallelism Strategy\end{tabular}} & \multicolumn{3}{c|}{Partitioning Dimension} & \multirow{2}{*}{\begin{tabular}[c]{@{}c@{}}Communication\\ Operation\end{tabular}} & \multirow{2}{*}{\begin{tabular}[c]{@{}c@{}}Communication\\ Volume\end{tabular}} & \multirow{2}{*}{\begin{tabular}[c]{@{}c@{}}Attention \\Parameters\end{tabular}}\\
            \cline{2-4}
            & $ \text{Linear}_{\text{QKV}} $ & Attention & $ \text{Linear}_{\text{Out}} $ & & & \\
            \hline
            Tensor Parallelism (TP)~\cite{shoeybi2019megatron} & H & H & H & All-Gather+Reduce-Scatter & $2bsh(n-1)/n$ & Partitioned \\
            Ulysses Sequence Parallelism (SP)~\cite{jacobs2023deepspeed} & S & H & S & All-to-All QKV\&Output & $2bsh(n-1)/n \times (2+2/m)/n$ & Replicated \\
            Context Parallelism (CP)~\cite{context_parallelism} & S & S & S & All-Gather KV & $2bsh(n-1)/n \times 2/m$ & Replicated \\
            \hline
            \hline
        \end{tabular}
    }
    % \vspace{-0.1in}
    \caption{Different parallelism strategies for self-attention. "H" denotes partitioning along the dimension of hidden size, while "S" denotes partitioning along the dimension of sequence length.}
    \vspace{-0.1in}
    \label{tab:background:parallelism}
\end{table*}
\fi

% \input{sections/overview}
\section{Communication-Efficient Parallelism}
\label{sec:design:parallelism}

With the rise of MoE models and the evolution of hardware compute capabilities, communication overhead has become increasingly critical in MoE training in production. In this section, we delve into the parallelism strategies employed to reduce communication volume and meet other training requirements, such as high GEMM (General Matrix Multiplication) efficiency.

\begin{figure}[t!]
    \centering
    \includegraphics[width=\linewidth]{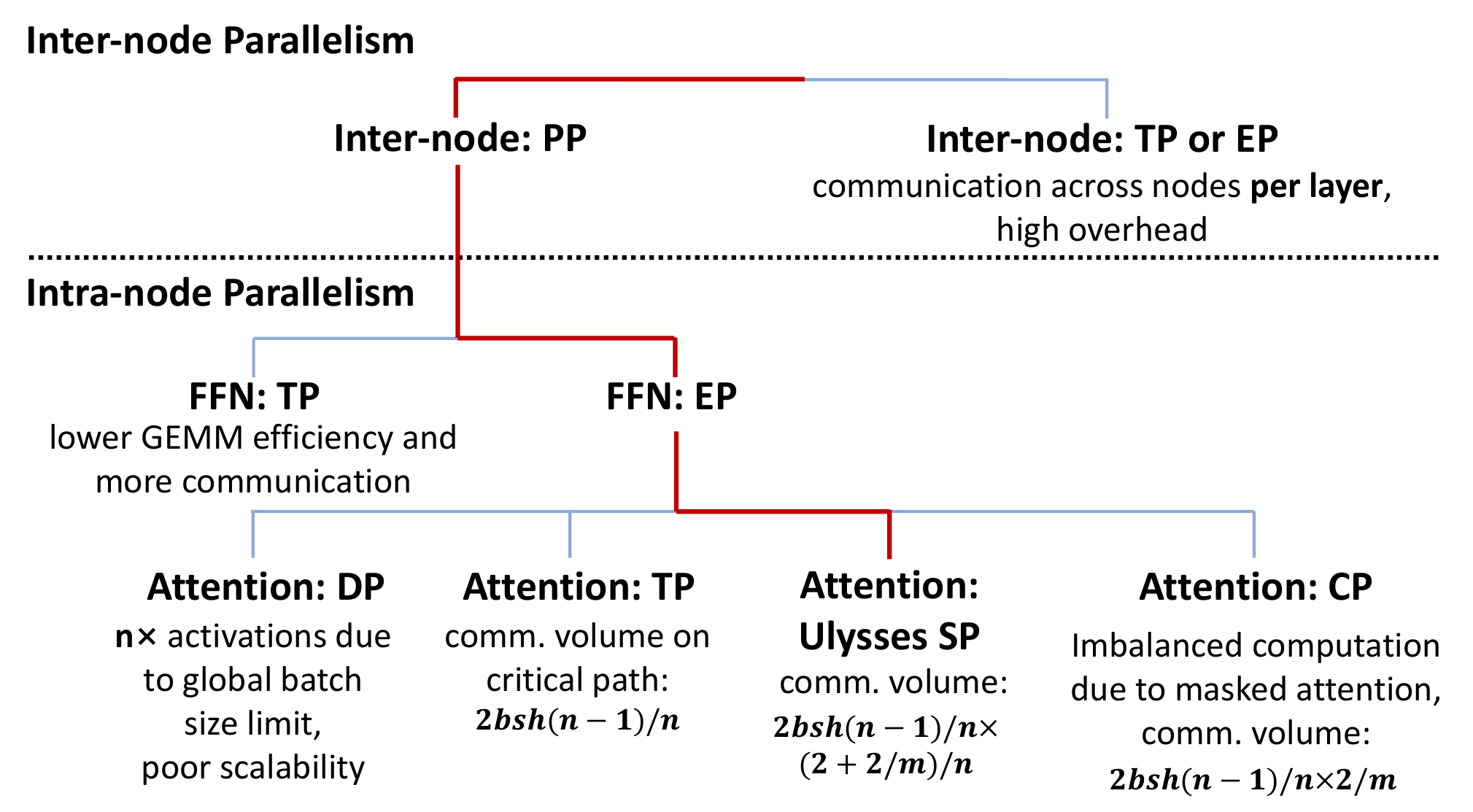}
    \vspace{-0.3in}
    \caption{Design space for large-scale MoE training.}
    \vspace{-0.1in}
    \label{fig:parallelism_design_space}
\end{figure}

Figure~\ref{fig:parallelism_design_space} shows the design space of parallelism strategies for large-scale MoE training, excluding the outermost data parallelism. We start with inter-node parallelism. Expert parallelism alleviates memory pressure from MoE models' large parameter size by distributing experts across nodes but incurs per-layer cross-node communication, harming training efficiency. Similarly, tensor parallelism's high communication overhead makes it more efficient to limit TP to a single node. Following prior work~\cite{jiang2024megascale}, we adopt pipeline parallelism to distribute model parameters, reduce communication, and overlap communication of different micro-batches.

Prior large-scale MoE training systems, such as Megatron-LM~\cite{shoeybi2019megatron} and DeepSpeed-MoE~\cite{rajbhandari2022deepspeed}, incorporate tensor parallelism to scale up training by partitioning the model parameters within the node. However, in our practice, we observe two issues with this approach: (1) TP partitions the expert dimension, which negatively impacts GEMM efficiency; and (2) TP introduces significant communication overhead, which remains constant as the parallelism size increases, eventually causing communication to exceed computation on modern hardware.

To address these issues, we tailor parallelism strategies for MoE model components. For feed-forward networks (i.e., experts), we replace tensor parallelism with expert parallelism and use custom communication modes optimized for varying top-k and expert sizes, ensuring communication overhead stays lower than tensor parallelism. For other components, we apply sequence parallelism, partitioning along the sequence dimension instead of the batch dimension, allowing scaling without increasing global batch size. This also reduces communication on critical paths compared to tensor parallelism. The additional memory and DP communication overhead remain manageable due to the parameter asymmetry across components. We detail the rationale and analysis of this intra-node parallelism strategy in the following sections. Table~\ref{tab:design:description} lists the key symbols.

% \begin{figure*}[t!]
%     \centering
%     \includegraphics[width=\linewidth]{figures/parallelism-comparison.pdf}
%     \newline
%     \hspace*{0.5em}
%         \begin{subfigure}{0.26\linewidth}
%             \caption{DP$\times$(DP$+$EP).}
%             \label{fig:parallelism-comparison:dp-dp-ep}
%         \end{subfigure}
%         % \hfill
%         \begin{subfigure}{0.33\linewidth}
%             \caption{EP$\times$(TP$+$TP).}
%             \label{fig:parallelism-comparison:ep-tp}
%         \end{subfigure}
%         % \hfill
%         \begin{subfigure}{0.35\linewidth}
%             \caption{\sysname: DP$\times$(SP$+$EP).}
%             \label{fig:parallelism-comparison:dp-sp-ep}
%         \end{subfigure}
%     \vspace{-0.1in}
%     \caption{Comparison of parallelism strategies. X$\times$(Y$+$Z) indicates that X represents the inter-node parallelism, while Y and Z denotes the intra-node parallelism for attention and FFN, respectively.}
%     \vspace{-0.1in}
%     \label{fig:parallelism-comparison}
% \end{figure*}

\subsection{Sequence Parallelism for Attention}
\label{sec:design:parallelism:sp_attention}

Due to the inherent parallelizability of the expert components in MoE models, most prior work on MoE training~\cite{rajbhandari2022deepspeed, li2023accelerating} focuses on optimizing expert parallelism, while data parallelism (DP) is typically applied to the non-MoE components such as attention. However, when scaling up MoE training, this approach proves insufficient due to the $n \times$ activation memory consumption. This issue arises because DP splits the batch dimension both across and within nodes. Compared to other intra-node parallelism strategies shown in Figure~\ref{fig:parallelism_design_space}, applying DP to attention forces each GPU within a node to process one micro-batch simultaneously, increasing the activation size by 8$\times$, which often results in out-of-memory issues.

\begin{table}[t!]
    \centering
    \resizebox{0.8\linewidth}{!} {
        \begin{tabular}{@{}ll@{}}
            \toprule
            Symbol & Description \\ \midrule
            $b$ & micro-batch size \\
            $s$ & sequence length \\
            $h$ & hidden dimension size \\
            $n$ & model parallelism (TP, SP, or EP) size \\
            $m$ & the ratio between the number of query heads and \\
            & that of key-value heads \\
            $k$ & number of experts that each token is routed to \\
            \bottomrule
        \end{tabular}
    }
    % \vspace{-0.1in}
    \caption{Description of symbols.}
    \vspace{-0.25in}
    \label{tab:design:description}
\end{table}

To enable scalable MoE training, implementing intra-node parallelism for the attention module is crucial. Tensor parallelism (TP) is commonly employed to parallelize attention operations within nodes. However, it introduces inevitable communication costs due to all-gathering and reduce-scattering activations along the critical path. With the increasing gap between computational FLOPs and communication bandwidth, we find that the TP communication overhead can even surpass the computation time of self-attention. This communication-dominated bottleneck limits the ability to overlap communication and computation, ultimately reducing training efficiency. 

We adopt sequence parallelism (SP), as proposed in DeepSpeed-Ulysses~\cite{jacobs2023deepspeed}, to scale MoE training and effectively reduce communication along the critical path. SP is commonly used in long-context training to address memory challenges associated with long inputs. We find it also works well in large-scale MoE training. First, it significantly reduces communication overhead compared to TP, especially when using grouped-query attention~\cite{ainslie2023gqa}. Second, while it introduces some parameter redundancy and increased communication overhead during parameter synchronization, the unique characteristics of MoE models make these trade-offs manageable and acceptable.

\parabf{Communication efficiency.} When utilizing TP, the communication volume in attention is
\begin{align}
    2bsh(n-1)/n. 
    \label{eq:1}
    % 2bsh(n-1)/n*2/m
\end{align}
With SP, the communication volume decreases to
\begin{align}
    2bsh(n-1)/n\times(2+2/m)/n, 
    \label{eq:2}
\end{align}
where $m$ represents the ratio between the number of query heads and that of key-value heads. Assuming the model is trained on an NVIDIA Hopper GPU workstation with an NVLink domain of size 8, the communication latency for sequence parallel attention can be reduced to about one-fourth of that required by tensor-parallel attention.

\begin{figure}
    \centering
    \includegraphics[width=0.85\linewidth]{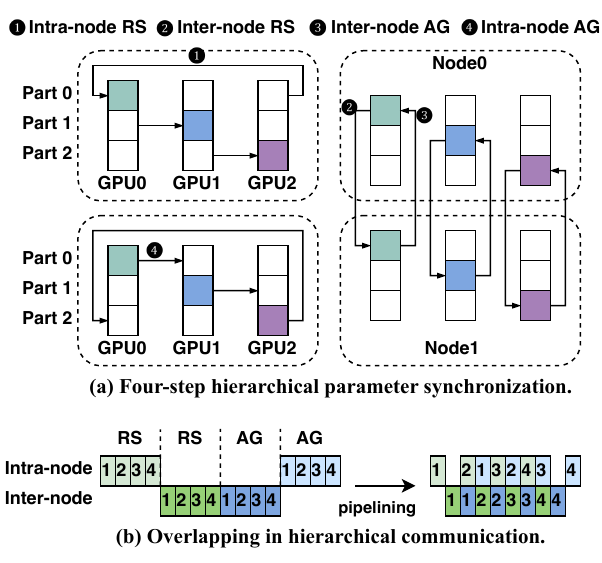}
    \vspace{-0.1in}
    \caption{Hierarchical communication for parameter synchronization in SP attention.}
    \vspace{-0.1in}
    \label{fig:hierarchy}
\end{figure}

\parabf{Data communication \& memory overhead.} A notable difference between SP and TP attention is how parameters are distributed across devices: TP shards the attention weights, while SP replicates them. This raises the concern about the potential increase in communication overhead for synchronizing gradients and parameters. Counterintuitively, given the intra- and inter-node bandwidth asymmetry and the adoption of hierarchical communication operations in modern communication libraries~\cite{nccl} as shown in Figure~\ref{fig:hierarchy} and analyzed in Appendix~\ref{apdx:hierarchical}, although SP attention requires synchronization of $n \times$ more parameters compared to TP attention, the difference in communication overhead is minimal in practical scenarios. 

% \begin{figure}
%     \centering
%     \begin{subfigure}[b]{\linewidth}
%         \centering
%         \includegraphics[width=0.99\textwidth]{figures/dp_comm_32gpu.pdf}
%         \label{fig:subfig1}
%     \end{subfigure}
%     \begin{subfigure}[b]{\linewidth}
%         \centering
%         \includegraphics[width=0.99\textwidth]{figures/dp_comm_64gpu.pdf}
%         \label{fig:subfig2}
%     \end{subfigure}
%      \vspace{-0.1in}
%     \caption{Latency vs. data size. A comparison of parameter synchronization latency between SP attention and TP attention with 32GPU and 64GPU.}
%      \vspace{-0.1in}
%     \label{fig:sp_dp_comm}
% \end{figure}

% In the experiment, we evaluated the communication latency of parameter synchronization between TP8 and SP8 across settings of 32 and 64 GPUs. We increase the data size from 384MB to 1536MB. The experiment results in figure \ref{fig:sp_dp_comm} demonstrate that the latencies for TP8 and SP8 are consistently comparable, with no notable differences observed. This observation corroborates our initial hypothesis that the two would exhibit similar performance characteristics in term of data parallelism communication latency.

On the other hand, the additional GPU memory consumption introduced by SP attention is minimal in MoE training. For large-scale MoE models with tens to hundreds of experts, the majority of GPU memory is consumed by the expert parameters. Our experiments, detailed in \S\ref{sec:evaluation:ablation}, confirm that the extra parameter synchronization and memory overhead of SP attention remain manageable.
% For instance, in a 2.7T-parameter MoE model with 128 experts, the attention parameters only account for less than 1\% of the total parameters.

\parabf{Balanced vs. imbalanced.} In addition to the Ulysses-style SP attention, we also explored other forms, including context parallelism (CP)~\cite{context_parallelism}, which partitions all activations along the sequence dimension. CP attention, however, faces workload imbalance due to causal masking in attention, as each token only attends to previous tokens. To mitigate this, we attempted the zigzag strategy by grouping the head and tail partitions of the sequence on the same GPU, although achieving perfect balance remains challenging. Consequently, in large-scale training, the entire training process is often constrained by the most imbalanced data batch. Moreover, this imbalance disturbs the training pipeline, thereby reducing overall training efficiency.

\subsection{Expert Parallelism for Feed-forward Network}
\label{sec:design:parallelism:ep}

% \begin{figure}[t!]
%     \centering
%     \includegraphics[width=0.5\textwidth]{figures/megascale-moe-parallelism.pdf}
%     % \vspace{-0.1in}
%     \caption{Parallelism.}
%     % \vspace{-0.1in}
%     \label{fig:parallelism}
% \end{figure}

\begin{figure}[t!]
    \centering
    \includegraphics[width=\linewidth]{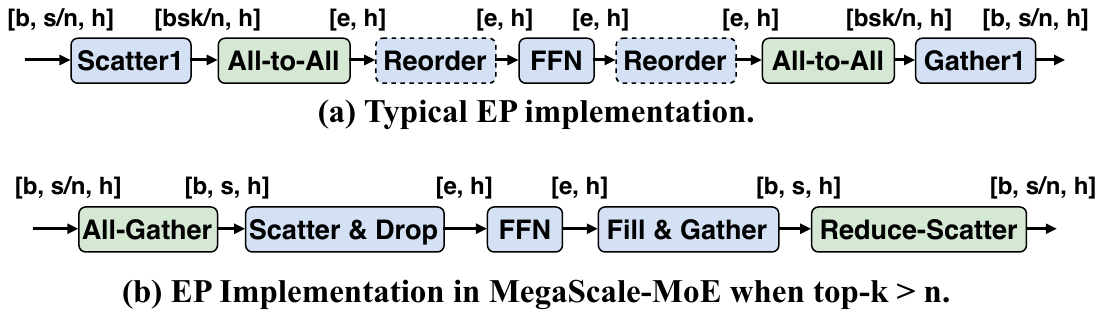}
    \vspace{-0.3in}
    \caption{Communication-efficient expert parallelism. $e$ represents the number of tokens routed to the worker.}
    \vspace{-0.1in}
    \label{fig:ep_impl}
\end{figure}

In the choice of parallelism strategies for the feed-forward network component, expert parallelism (EP) consistently outperforms tensor parallelism. TP partitions the hidden dimension of each expert, reducing GEMM efficiency, whereas EP maintains full expert computation on each device. Theoretically, the communication cost for EP is 
\begin{align}
    2k/n\times bsh(n-1)/n, \label{eq:3}
\end{align}
while for TP it is
\begin{align}
    2bsh(n-1)/n. \label{eq:4}
\end{align}
Although their relative efficiency depends on the ratio $k/n$, we design an adaptive communication strategy for different top-$k$ values to minimize the communication volume of EP.

\begin{figure}[t!]
    \centering
    \includegraphics[width=0.7\linewidth]{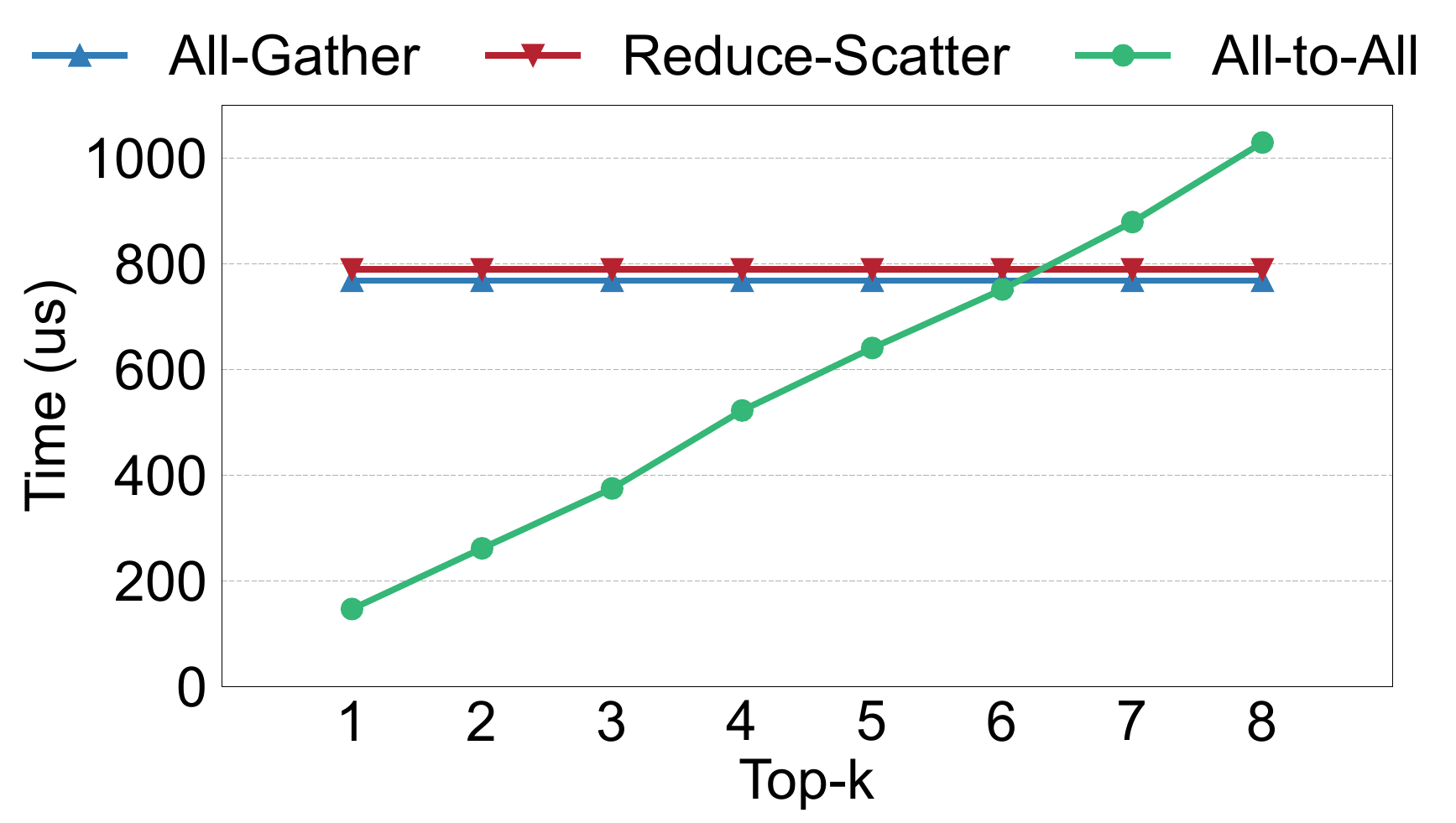}
    \vspace{-0.1in}
    \caption{Comparison of AG, RS, and A2A for token dispatch.}
    \vspace{-0.1in}
    \label{fig:ag-rs-a2a}
\end{figure}

\parabf{Efficient communication pattern.} Figure~\ref{fig:ep_impl} compares the typical EP implementation with \sysname's approach. The standard EP implementation requires two all-to-all communications for token dispatch and aggregation. Additionally, a scatter operation may be required before sending and after receiving tokens to ensure that tokens assigned to the same expert reside in a contiguous memory space. 

When the top-$k$ value exceeds $n$, we replace traditional all-to-all communication with all-gather and reduce-scatter. First, an all-gather operation collects tokens from all workers. Then, a local scatter operation discards unneeded tokens, retaining only those required by the experts on the current worker. After expert computation, the tokens are assembled into a complete tensor. This approach enables a gather operation before communication, followed by a reduce-scatter to produce the final result, ensuring that EP’s communication overhead remains equal to or lower than TP’s.

In practical training, all-to-all communication is less efficient than all-gather and reduce-scatter, as it requires each worker to communicate with all others, whereas all-gather and reduce-scatter follow a ring-based communication pattern with only neighboring workers. As shown in Figure~\ref{fig:ag-rs-a2a}, the communication time for these three operations in Mixtral-8$\times$7B reveals that when top-$k$ > 6, the all-gather-based EP implementation is more efficient.

% In typical EP implementations, tokens are dispatched to different devices based on the selected experts using all-to-all communication. After completing the expert computations, tokens are sent back to the original device using another all-to-all communication for subsequent processing.

% In common case with $topk$ equals 1, as there is no need for weighted gathering among scattered tokens, our observation suggests that reverting all tokens to their original positions after the FFN computation may not be necessary. Instead, tokens can remain on their assigned devices for the attention computations of the following layer, and during the all-to-all communication step in sequence-parallel attention, they can be transferred to the correct positions as needed. This approach eliminates one all-to-all communication in expert parallelism, thereby reducing the communication volume by half.

% \parabf{Computation efficiency.} Beyond reducing communication volume, EP enhances the computational efficiency of the feed-forward network component. Traditional tensor parallelism partitions the dimension of expert GEMM operations, and as these dimensions are divided more finely, computational efficiency decreases. In contrast, EP divides the number of expert GEMM operations on the worker, which maintains the inherent efficiency of each GEMM operation.

\parabf{Efficient operators.} Instead of using \texttt{torch.scatter\_add} and \texttt{torch.gather} for tensor scattering and gathering like Megatron-LM, we develop efficient scatter and gather operators directly using CUDA. Based on the token routing results, we pre-calculate the mapping from each row of the input tensor (representing a token) to the corresponding row in the output tensor. The scatter and gather operators then perform data transfers efficiently according to this mapping. 

\parabf{Load balance.} A well-known challenge in MoE model training is load balancing across experts~\cite{li2023accelerating, liu2024deepseek}. To address this, we use auxiliary loss and token dropping to balance the workload across GPUs within each node. Similar to DeepSeek-V2~\cite{liu2024deepseek}, we treat the experts placed on the same GPU as a group and calculate the balance loss and computational capacity for each device rather than for each individual expert. 

\begin{figure}[t!]
    \centering
    \includegraphics[width=0.9\linewidth]{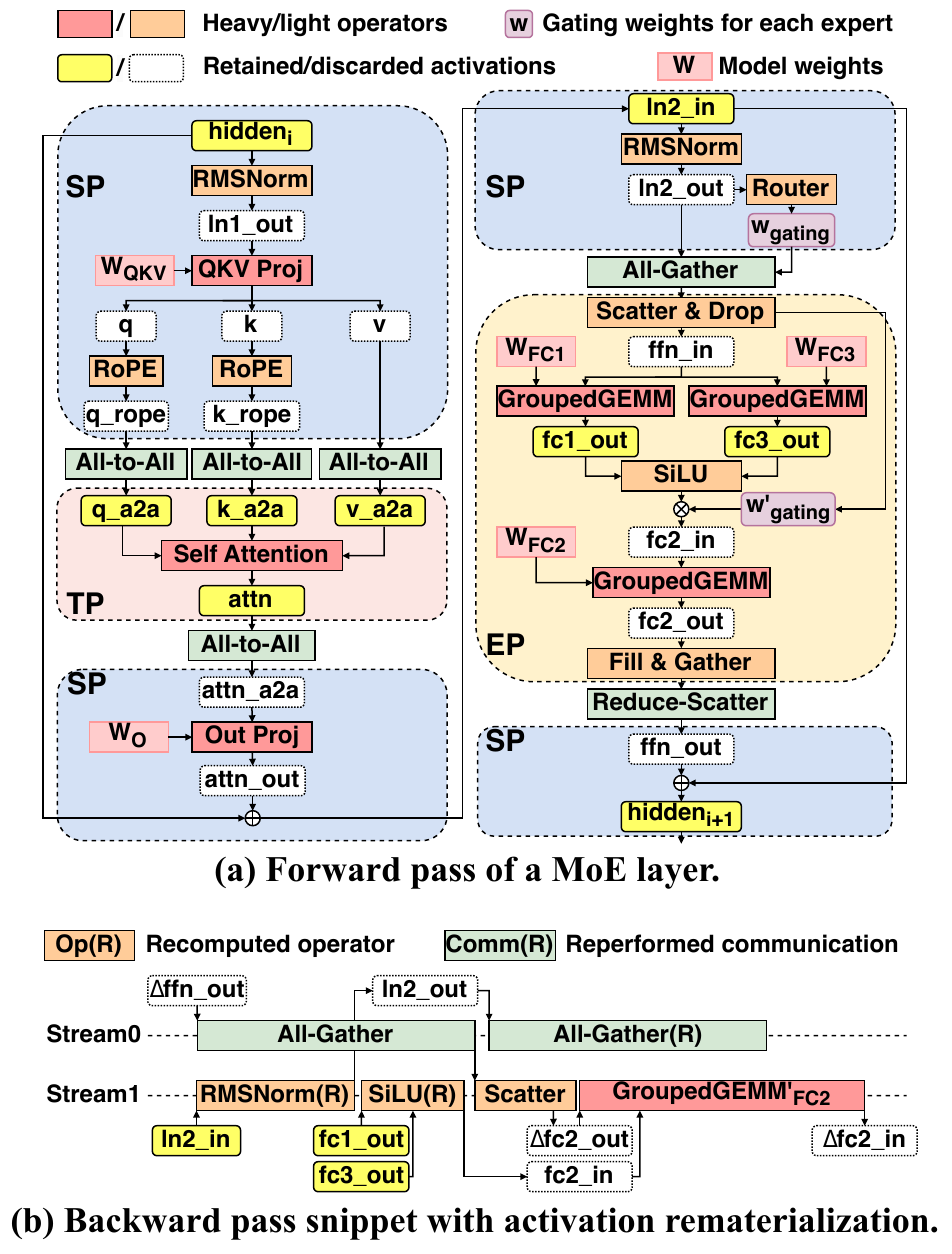}
    \vspace{-0.15in}
    \caption{Selective activation rematerialization.}
    \vspace{-0.15in}
    \label{fig:moe_computation}
\end{figure}

\section{Communication-computation Overlap}
\label{sec:design:overlapping}

After optimizing parallelism strategies to minimize communication volume, we further reduce the communication overhead to nearly zero using comprehensive communication-computation overlapping techniques. Training large models involves integrating various techniques, which increases the complexity of communication overlap. For instance, at any given moment, the device might concurrently handle computation and communication kernels, overlap PP and DP communications, and manage data transfers between the device and host. 
Existing frameworks like Megatron-LM assemble attention and FFN modules into MoE layers and rely on the \texttt{torch.autograd} package for backward propagation, which limits the flexibility of communication overlap. In contrast, \sysname decomposes the attention and FFN modules of each MoE layer into operators that run as GPU kernels, enabling fine-grained communication overlap through flexible scheduling.

\subsection{Inter-operator Overlap}
\label{sec:design:overlapping:inter-op}

We overlap communication operators with independent computation operators by executing them asynchronously on different CUDA streams. To achieve optimal performance during the training process, we adopt a specifically hand-tailored, holistic scheduling strategy. 
% Since performance is critical, we adopted a specifically hand-tailored, comprehensive holistic scheduling strategy to ensure optimal performance across the whole training process.

\parabf{Holistic scheduling.} From the caller’s perspective, we implement a unified macro module to execute the entire MoE layer’s forward and backward passes, thereby expanding our scheduling flexibility. For instance, during the backward pass, various communication operators can be overlapped with dependency-free computations, such as activation recomputation, to improve efficiency. From the runtime perspective, a key challenge is efficiently managing concurrent communication tasks by resolving resource conflicts to prevent blocking and maximize throughput. This requires careful coordination, such as determining the number of SMs allocated to each communication operator, to minimize interference and optimize overall throughput.

\begin{figure}[t!]
    \centering
    \includegraphics[width=0.5\textwidth]{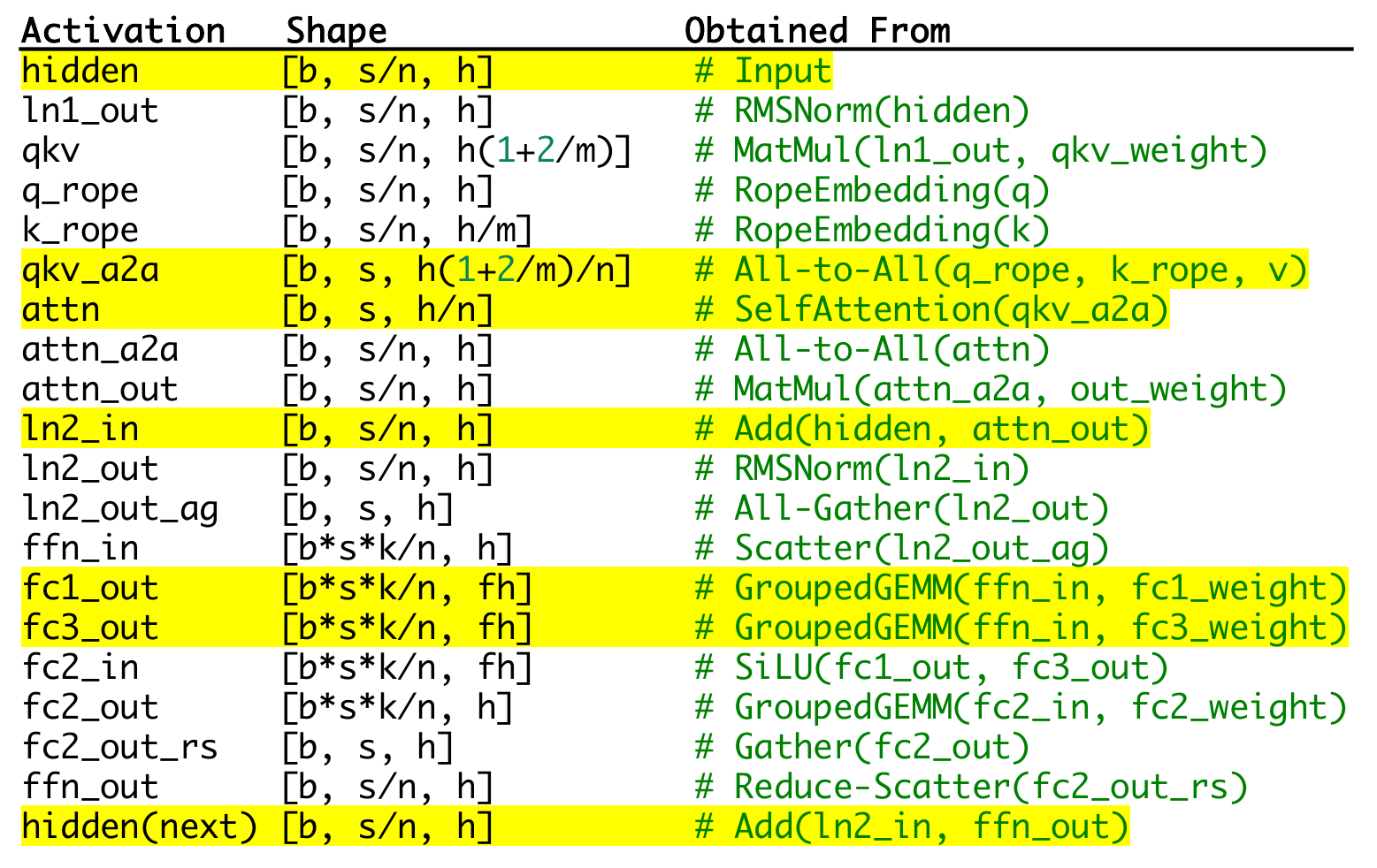}
    \vspace{-0.3in}
    \caption{Activation shapes in rematerialization.}
    \vspace{-0.1in}
    \label{fig:memory}
\end{figure}

\begin{figure*}[t!]
    \centering
    \includegraphics[width=\linewidth]{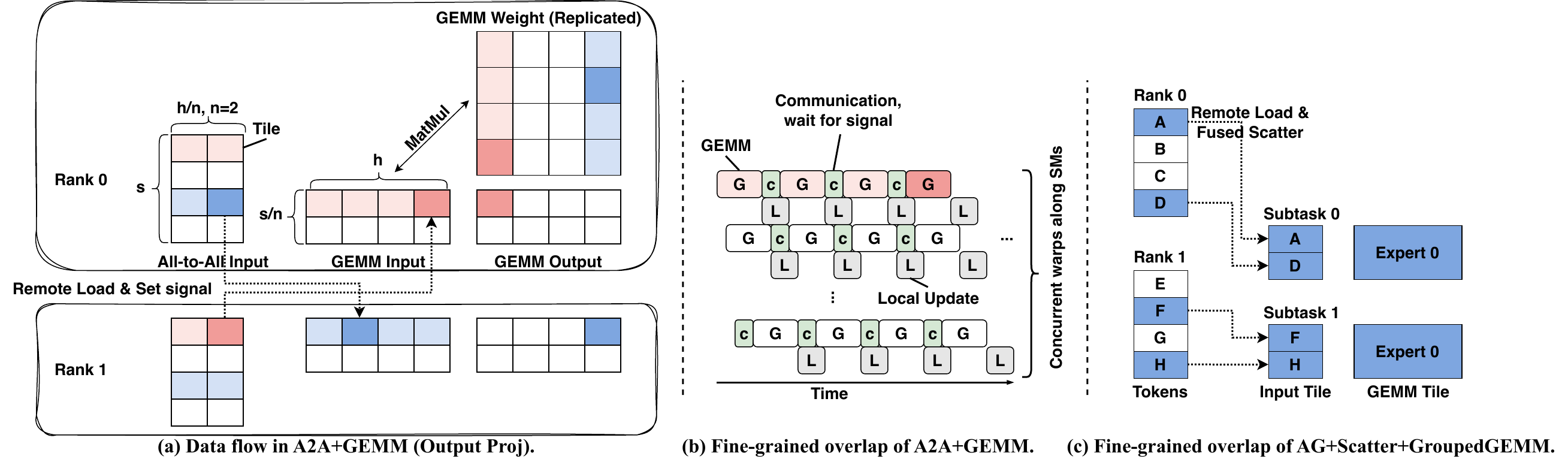}
    \vspace{-0.2in}
    \caption{Fine-grained intra-operator communication-computation overlap.}
    \vspace{0.1in}
    \label{fig:fine-grained}
\end{figure*}

\parabf{Selective activation rematerialization.} The holistic scheduling strategy also helps reduce memory usage without compromising training speed. Compared to dense models with equivalent computational requirements, MoE models exert significantly higher memory pressure during training due to their parameter count being several times larger. In addition to employing ZeRO optimizations~\cite{rajbhandari2020zero} to eliminate redundant optimizer states across DP groups, we further optimize memory usage through selective activation rematerialization. This approach reduces activation memory requirements by re-performing computation and communication operators that can be overlapped with other necessary operators.

Figure~\ref{fig:moe_computation}a illustrates the forward pass of a Mixtral~\cite{jiang2024mixtral} MoE layer and highlights key activations produced during this process. \sysname strategically retains activations that are computationally expensive to recompute, while recalculating others generated by memory-intensive operations or communication operations. This minimizes dependencies on backward computation, enabling rematerialization operations to overlap with other computations and communications, avoiding delays in the critical path. 
For example, as shown in Figure~\ref{fig:moe_computation}b, the backward pass of the GroupedGEMM operator for FC2 requires the activation \texttt{fc2\_in} and the gradient of  \texttt{fc2\_out} (denoted as \texttt{$\Delta$fc2\_out}) as inputs. \sysname recomputes \texttt{fc2\_in} and overlaps this operator with gradient communication (i.e., all-gather for \texttt{$\Delta$ffn\_out}). Similarly, \texttt{ffn\_in} is obtained through re-performing \texttt{RMSNorm} and all-gather, with these operators hidden within the preceding communication and the FC2 GroupedGEMM, respectively. 
\sysname also places the weighted sum of \texttt{ffn\_out} immediately after the SwiGLU~\cite{shazeer2020glu} activation function to eliminate the need to store \texttt{ffn\_out}. This reordering ensures computational consistency by avoiding operators that cross non-linear boundaries.

Figure~\ref{fig:memory} illustrates the shapes of the key activations produced during forward propagation, with the highlighted activations retained for backward propagation.
Let the model parallelism size within one MoE layer be $n$ and the intermediate hidden size of one expert be $fh$. The total activation of a single MoE layer is 
\begin{align}
% (4K+3K*FH/H+12+5/mqa)BSH/n
(2n+2k+3kf+12+5/m)bsh/n,
\nonumber
\end{align}
which we have reduced to 
\begin{align}
(2kf+4+2/m)bsh/n.
\nonumber
\end{align}
\sysname reduces the activation memory by $\sim 50\%$ while maintaining the same training speed.
 
\subsection{Intra-operator Overlap}
\label{sec:design:overlapping:intra-op}

Although inter-operator overlap effectively hides communication latency, squeezing all bubbles in the execution timeline remains non-trivial---especially in the forward pass, where no rematerialization or gradient computation operators exist to overlap with communication. Some forward operators directly depend on communication, such as token dispatch for expert computation, making overlap impossible unless another micro-batch is introduced, which increases memory pressure.

A widely adopted solution~\cite{jiang2024megascale, transformer-engine, wang2022overlap} is to decompose operators into smaller parallel ones to enable pipelining by executing them on separate CUDA streams. 
However, this approach introduces non-negligible overhead: $(i)$ complex stream control, involving host interference and causing random bubbles due to the non-deterministic feature of CPU control; $(ii)$ imperfect tail computation, increasing overall computation latency.
% The reason behind this is twofold. First, some computation operators have a direct dependency on communication operators, making it impossible to hide latency between them unless we introduce another micro-batch (which will increase memory pressure). Second, inter-operator overlap requires complex stream control, which involves host interference and may cause random bubbles due to the non-deterministic feature of CPU control.

To address the above issues, we adopt intra-operator overlap to parallelize communication and computation operators with direct dependencies. The core idea is to fuse these operators and break down the workloads into tiles.
Following prior work~\cite{jangda2022breaking, chang2024flux, zhang2025comet, zheng2025tilelink}, we implement barriers in device memory between communication and computation operators. 
These barriers enable fine-grained tile-level notifications and remove the need for host interference, further improving training performance.
We implement two types of kernels, overlapping with GEMMs and overlapping with MoE GroupedGEMMs, for the attention and FFN modules, respectively.

\parabf{Overlapping with GEMMs.} We first introduce the intra-operator communication-computation overlap for GEMM kernels.
Specifically, we implement all-to-all(A2A)+GEMM and GEMM+A2A kernels for Output and QKV Projections in SP attention, respectively, 
where X+Y means Y executed after X.
Figure~\ref{fig:fine-grained} shows the data flow and overlapping pattern in A2A+GEMM. 
The GEMM on local data and communication for remote data starts simultaneously. We leverage dedicated GPU copy engines for data transfer, ensuring that all SMs (streaming multiprocessors) are fully utilized for computation. Once a remote data tile arrives at local memory, a signal notifies the GEMM kernel to continue its computation on the arrived tile.
For GEMM+A2A, the all-to-all operation is fused into the GEMM kernel. Each tile of GEMM computation ends with a remote data transfer that writes the output data tile to remote ranks. 
We also implement all-gather+GEMM and GEMM+reduce-scatter kernels for tensor parallelism, which are similar to A2A+GEMM and GEMM+A2A.

For A2A+GEMM and GEMM+A2A, we allocate a small number of SMs for communication as all-to-all is more complex than all-gather and reduce-scatter. The number of SMs for communication is tuned to make communication and computation exhibit similar latency.
Moreover, multiple ranks may simultaneously read from or write to the same device, potentially causing contention in NVLink. To mitigate this, we apply swizzling~\cite{chang2024flux, zhang2025comet, zheng2025tilelink} to reorder tile communication and computation so that the arrival of communication tiles aligns with the pace of computation tiles.

\parabf{Overlapping with GroupedGEMMs} 
For expert parallelism with token dispatch and combine, we aim to overlap communication with GroupedGEMMs. We implement two types of overlapping kernels: all-gather+scatter+GroupedGEMM and GroupedGEMM+gather+reduce-scatter. Unlike the overlapping techniques for GEMM kernels, MoE GroupedGEMMs require token shuffling (scatter/gather). As a result, each computation tile may depend on tokens from multiple ranks. To effectively overlap computation with communication, we sort the token order to minimize the number of dependent ranks for each computation tile. Additionally, since each tile has its own dependencies, the signal control for each tile varies depending on the MoE routing, which is determined dynamically.
In detail, for AG+scatter+GroupedGEMM, we reorder tokens along the sequence dimension based on their routed expert index. Then, for each expert, we sort the routed tokens according to their source rank index. Finally, we slice the sorted sequence into blocks and perform GroupedGEMM using a sequence of computation tiles. Specifically, as shown in Figure~\ref{fig:fine-grained}c, we fuse the local scatter into the kernel by selecting rows of input data based on the index mapping. The GroupedGEMM computation for each expert is divided into tiles, with each tile depending on only a subset or even a single source rank. 
This reduces the overall waiting time for each computation block, avoids redundant loading of expert parameters, and improves the overlap between computation and communication tiles.

% We primarily rely on the CUTLASS programming template~\cite{cutlass} to generate highly efficient GEMM kernels, and integrate NVSHMEM~\cite{nvshmem} within kernels for fine-grained communication. Unlike NCCL~\cite{nccl}, which provides a higher-level abstraction for comprehensive communication operations, NVSHMEM offers a more composable, low-level API, allowing for more granular control over data access within kernels.

\section{Communication Compression}
\label{sec:design:precision}

We further reduce communication overhead by applying communication compression. To maintain convergence stability, mixed-precision training frameworks typically transfer tensors awaiting reduction in higher precision, such as FP32, to ensure more accurate accumulation. A common example of this is gradient reduce-scatter in data parallelism.

\parabf{DP communication compression.} As MoE model parameters increase, so does the communication overhead for parameter and gradient synchronization in data parallelism. Prior work has explored gradient compression to mitigate this cost. In our BF16 mixed-precision training, we carefully apply FP32-to-BF16 precision reduction for gradient synchronization, balancing efficiency and convergence stability.

\begin{figure}[t!]
    \centering
    \includegraphics[width=0.9\linewidth]{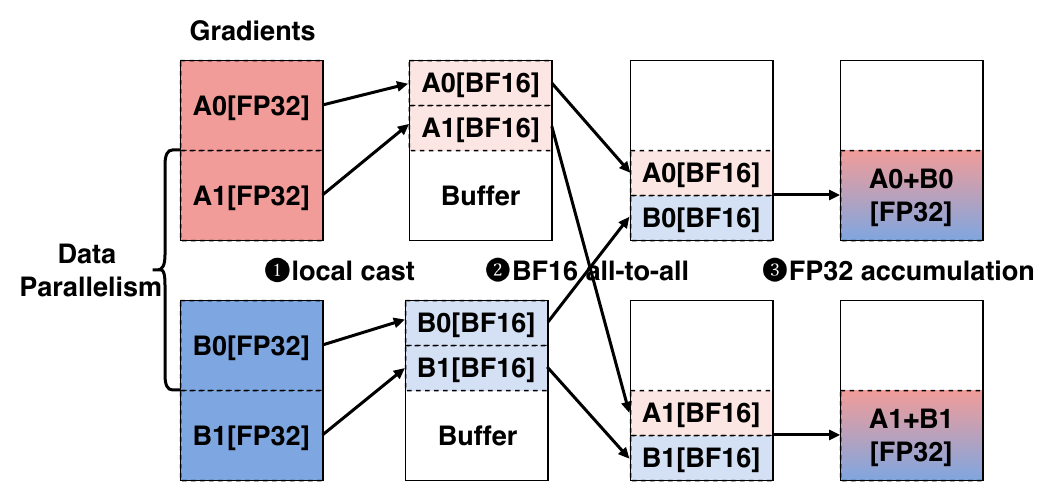}
    \vspace{-0.1in}
    \caption{DP communication compression.}
    \vspace{-0.1in}
    \label{fig:compression}
\end{figure}

Specifically, as shown in Figure~\ref{fig:compression}, we retain the main gradients in FP32 during local gradient accumulation in pipeline parallelism. After each model stage completes accumulation, instead of relying solely on reduce-scatter for gradient synchronization, we cast gradients to BF16 and perform all-to-all communication within the data parallel group to gather the required gradient shards, which are then locally aggregated in FP32. Our results show that this approach introduces negligible precision loss compared to directly performing reduce-scatter with FP32, while reducing gradient communication overhead by 50\%.

This approach minimizes risk for two key reasons. First, it performs a one-time conversion of accumulated gradients to BF16 during communication, while the local gradient accumulation is maintained in FP32 precision. Second, instead of using ring-style reduce for BF16 gradient communication, it employs all-to-all communication, with the final reduction computed using FP32 summation. This design prevents precision loss that could arise from repeated accumulation of BF16 values in ring-based reductions.

We observe that casting large gradients and performing all-to-all communication increases peak memory consumption, potentially causing out-of-memory errors. To mitigate this, we develop a memory-efficient operator that in-places BF16 gradients into half of the FP32 input buffer while using the remaining half as the output buffer for BF16 all-to-all communication, preventing peak memory growth.

\begin{table}[t!]
    \centering
    \resizebox{\linewidth}{!} {
        \begin{tabular}{@{}cccccccc@{}}
            \toprule
            Name & \#layers & $h$ & \#heads & $m$ & $h_{ffn}$ & \#experts & top-$k$ \\\midrule
            % \multirow{4}{*}{\begin{tabular}[c]{@{}c@{}}Open-\\Source\end{tabular}} & 
            Internal-352B & 60 & 4096 & 32 & 4 & 14336 & 32 & 3 \\
            % Internal-1T & 100 & 8192 & 64 & 8 & 12288 & 32 & 3 \\
            Mixtral-8$\times$7B & 32 & 4096 & 32 & 4 & 14336 & 8 & 2 \\
            Mixtral-8$\times$22B & 56 & 6144 & 48 & 6 & 16384 & 8 & 2 \\
            Hunyuan-Large & 64 & 6400 & 80 & 10 & 18304 & 16 & 1 \\
            Phi-3.5-MoE & 32 & 4096 & 32 & 4 & 6400 & 16 & 2 \\
            DeepSeekMoE & 28 & 2048 & 16 & 1 & 1408 & 64 & 6 \\
            \bottomrule
        \end{tabular}
    }
    % \vspace{-0.1in}
    \caption{Model configurations in evaluation.}
    \vspace{-0.2in}
    \label{tab:evaluation:model_conf}
\end{table}

\parabf{Communication compression for FP8 training.} In low-precision FP8 training, the proportion of communication time increases due to reduced computation time. To mitigate communication overhead, we explore compressing communication volume using FP8 precision with appropriate quantization techniques. 
Currently, we apply communication compression in FP8 MoE training with tensor parallelism, focusing on reduction scenarios prone to overflow or underflow. For example, we adopt the E4M3 format (4-bit exponent and 3-bit mantissa) for all tensors. Similar to DP reduce-scatter compression, we replace BF16 TP reduce-scatter with FP8 all-to-all in forward propagation and perform reduction in FP32 precision. In the corresponding backward propagation, we apply FP8 all-gather for gradients. Notably, simply reducing precision leads to loss misalignment with BF16 training. To mitigate this, we apply per-token activation quantization for forward communication and per-channel quantization for backward communication. In backward propagation, we further group quantization along the token dimension using a small group size (e.g., 128).

\section{Evaluation}
\label{sec:evaluation}

In this section, we present a comprehensive evaluation of \sysname, covering overall training performance (\S\ref{sec:evaluation:performance}), ablation studies of \sysname's key optimizations (\S\ref{sec:evaluation:ablation}), and the effectiveness of the precision-communication co-design (\S\ref{sec:evaluation:co-design}). Table~\ref{tab:evaluation:model_conf} lists the configurations of the MoE models used in our evaluation, detailing hidden size ($h$), FFN intermediate size ($h_{ffn}$), number of experts, and top-$k$ values. 
The evaluation is conducted on NVIDIA H800 GPUs unless otherwise specified, with the specifications provided in Table~\ref{tab:evaluation:gpu_spec}. 

\begin{table}[t!]
    \centering
    \resizebox{\linewidth}{!} {
        \begin{tabular}{c|c|c|c|c}
            \hline
            \hline
            \multirow{2}{*}{\begin{tabular}[c]{@{}c@{}}System\end{tabular}} & \multirow{2}{*}{\begin{tabular}[c]{@{}c@{}}\#GPUs\end{tabular}} & \multirow{2}{*}{\begin{tabular}[c]{@{}c@{}}Iteration\\ Time (s)\end{tabular}} & \multirow{2}{*}{\begin{tabular}[c]{@{}c@{}}Throughput\\ (tokens/s)\end{tabular}} & \multirow{2}{*}{\begin{tabular}[c]{@{}c@{}}Training Time for\\ 1T Tokens (days)\end{tabular}} \\
            & & & & \\\cline{1-5}
            \multirow{5}{*}{Megatron-LM} & 240 & 39.94 & 151.1k & 76.61 \\
            & 480 & 19.56 & 301.1k & 38.38 \\
            & 720 & 13.70 & 430.5k & 26.88 \\
            & 960 & 10.82 & 550.2k & 21.23 \\
            & 1440 & 7.90 & 746.6k & 15.50 \\\cline{1-5}
            \multirow{5}{*}{\sysname} & 240 & 21.61 & 272.9k (\textbf{1.81$\times$}) & 42.41 \\
            & 480 & 11.83 & 498.6k (\textbf{1.65$\times$}) & 23.21 \\
            & 720 & 7.97 & 740.1k (\textbf{1.72$\times$}) & 15.64 \\
            & 960 & 6.12 & 963.8k (\textbf{1.77$\times$}) & 12.01 \\
            & 1440 & 4.19 & 1407.7k (\textbf{1.88$\times$}) & 8.22 \\
            \hline
            \hline
        \end{tabular}
    }
    % \vspace{-0.1in}
    \caption{Strong-scaling training performance for the 352B MoE model with NVIDIA H800 GPUs. The number in parentheses in the throughput column represents the speedup of \sysname compared to Megatron-LM.}
    \vspace{-0.2in}
    \label{tab:evaluation:strong-scaling}
\end{table}

\subsection{Training Performance}
\label{sec:evaluation:performance}

\sysname is built on top of Megatron-LM~\cite{shoeybi2019megatron}, a state-of-the-art open-source LLM training system that supports 3D parallelism strategies and is continuously updated to incorporate the latest optimizations from the community. Our evaluation uses the Megatron-LM on GitHub~\cite{megatron-lm} with commit hash f1f03922, selected for its stability at the commencement of our experiments months ago. For fair comparison, we use the same global batch size for Megatron-LM and \sysname and choose the optimal parallelism configurations for the two systems, respectively. Specifically, \sysname employs SP attention and EP within each node, while Megatron-LM adopts TP within each node, with both systems configured with a PP size of 15. \revision{We tune the configuration of Megatron-LM to meet its requirement of a uniform TP size across all components. As discussed in \S\ref{sec:design:parallelism:sp_attention}, for Megatron-LM, a TP size of 1 leads to a prohibitive 8$\times$ activation memory (addressable only with slow recomputation via gradient checkpointing), while a TP size of 8 forces EP to operate across nodes, incurring more communication costs than PP. Notably, both systems in the evaluation enable the communication-computation overlap techniques from MegaScale~\cite{jiang2024megascale} for data and pipeline parallelism. Therefore, the communication overhead mainly comes from intra-node model parallelism, e.g. TP, SP and EP.} Sequence length is 8,192 and vocabulary size is 65,536. 

\begin{figure}[t!]
    \centering
    \includegraphics[width=0.8\linewidth]{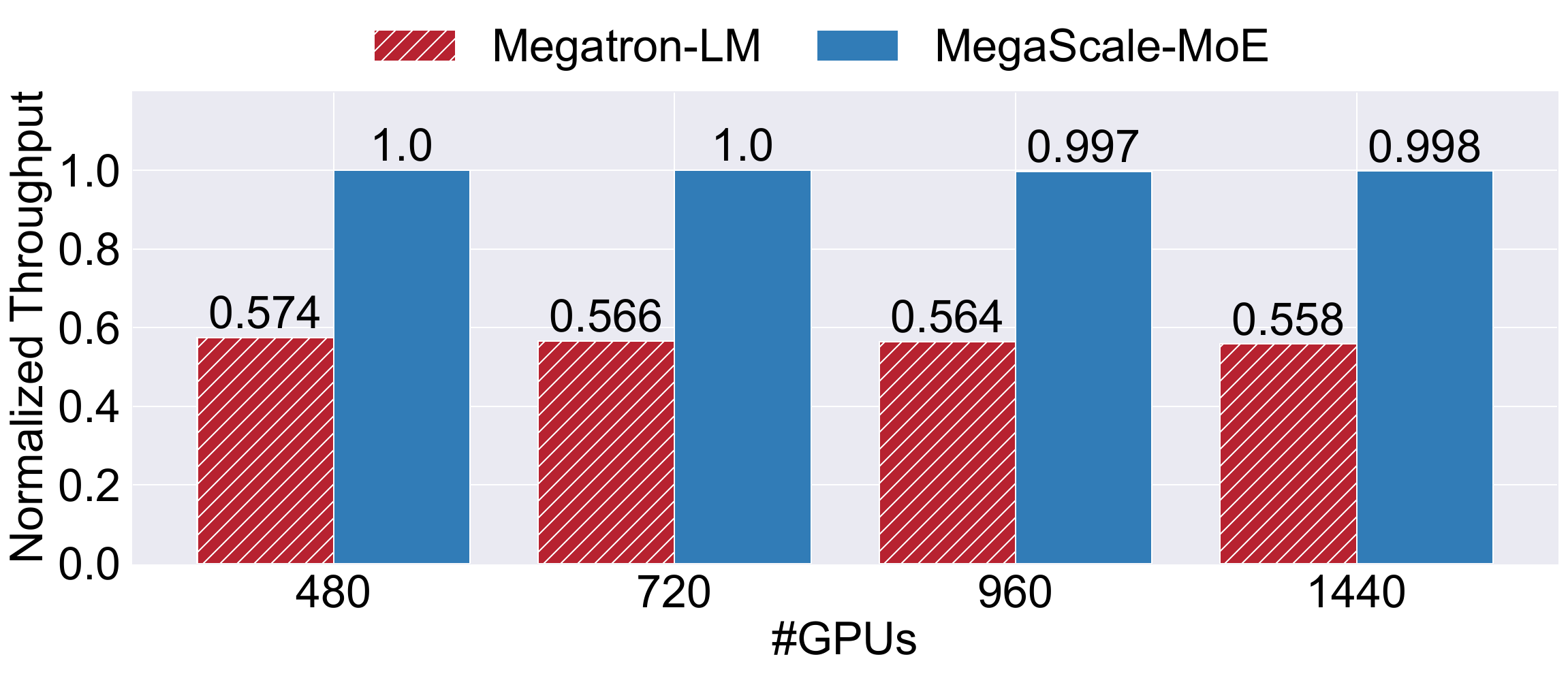}
    \vspace{-0.15in}
    \caption{Weak-scaling training performance for the 352B MoE model with NVIDIA H800 GPUs.}
    \vspace{-0.1in}
    \label{fig:eval:weak-scaling}
\end{figure}

\parabf{Scalability.} Table~\ref{tab:evaluation:strong-scaling} compares the strong-scaling training performance of Megatron-LM and \sysname on the 352B MoE model. We scale the number of GPUs while keeping the global batch size fixed at 720. 
% This experimental setting is more realistic, given that batch size is constrained by convergence effects and cannot be indefinitely scaled with the number of GPUs. 
Across all settings, \sysname achieves 1.65–1.88$\times$ speedups over Megatron-LM. As the number of GPUs increases, the MFU (Model FLOPs Utilization) of \sysname declines from 32.48\% to 27.89\%. This is expected, as the batch size is fixed and the number of micro-batches for each pipeline decreases with more GPUs, leading to more bubbles. 

Figure~\ref{fig:eval:weak-scaling} presents the weak-scaling training performance of Megatron-LM and \sysname on the same model. We scale the global batch size from 360 to 1,080 in proportion to the number of GPUs (from 480 to 1,440). \sysname achieves a 1.74-1.79$\times$ training throughput compared to Megatron-LM. As the scale increases, Megatron-LM’s throughput degrades by 2.74\% due to increased communication overhead. In contrast, \sysname exhibits near-linear scalability, with its throughput declining by only 0.2\%, benefiting from comprehensive communication-computation overlap. 
% \sysname achieves a 12.67\% higher MFU than Megatron-LM. As the scale increases, Megatron-LM’s MFU drops by 0.46\% due to increased communication overhead, whereas \sysname exhibits near-linear scalability, benefiting from comprehensive communication overlap. 

\begin{figure}[t!]
    \centering
    \includegraphics[width=\linewidth]{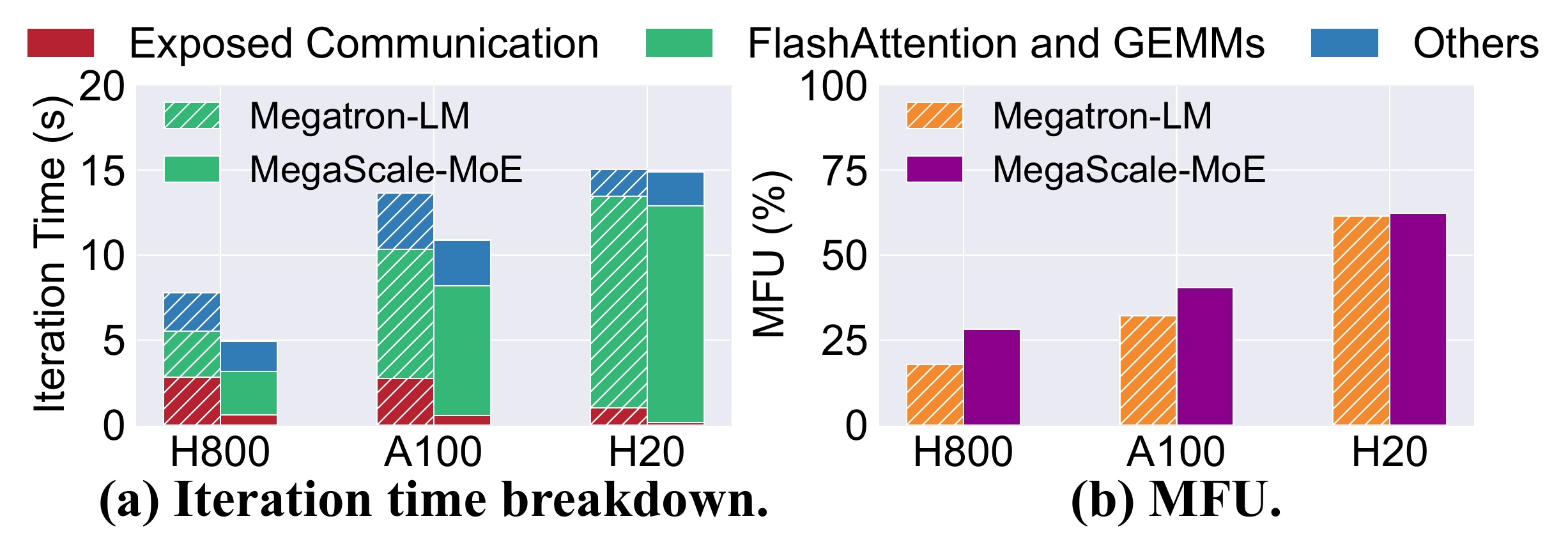}
    \vspace{-0.3in}
    \caption{Performance breakdown of training Mixtral-8$\times$7B on different GPUs.}
    \vspace{-0.1in}
    \label{fig:eval:different_gpus}
\end{figure}

\begin{table}[t!]
    \centering
    \resizebox{0.9\linewidth}{!} {
        \begin{tabular}{c|c|c|c|c}
            \hline
            % \hline
            \multirow{2}{*}{GPU} & 
            \multirow{2}{*}{\begin{tabular}[c]{@{}c@{}}Compute Cap-\\ability (TFLOPS)\end{tabular}} & 
            \multicolumn{2}{c|}{Memory Spec.} & 
            \multirow{2}{*}{\begin{tabular}[c]{@{}c@{}}NVLink\\ Bw. (GB/s)\end{tabular}} \\
            \cline{3-4}
            & & Cap. (GB) & Bw. (TB/s) & \\
            \hline
            H800 & 989 & 80 & 3.4 & 400 \\
            A100 & 312 & 80 & 2.0 & 600 \\
            H20 & 148 & 96 & 4.0 & 900 \\
            \hline
            % \hline
        \end{tabular}
    }
    % \vspace{-0.1in}
    \caption{Specifications of different NVIDIA GPUs.}
    \vspace{-0.15in}
    \label{tab:evaluation:gpu_spec}
\end{table}

\parabf{Performance breakdown on different GPUs.} We conduct a deep dive into \sysname to further understand the performance of training a MoE model in production environments. We train Mixtral-8$\times$7B on 32 NVIDIA H800, H20, and A100 GPUs, respectively. The specifications of GPUs we used are listed in Table~\ref{tab:evaluation:gpu_spec}. We set the DP size as four, the TP size as eight for Megatron-LM, and the SP and EP size as eight for \sysname. As shown in Figure~\ref{fig:eval:different_gpus}b, across the four kinds of GPUs, \sysname consistently outperforms Megatron-LM by up to 1.58$\times$ in MFU. 
Figure~\ref{fig:eval:different_gpus}a demonstrates the iteration time breakdown of Megatron-LM and \sysname. Exposed communication time represents the communication time that is not overlapped with computation operations. FlashAttention and GEMMs are the operations we count when calculating MFU. The performance gain primarily results from \sysname's communication-efficient parallelism strategies and fine-grained overlapped communication.

Note that the MFU value decreases as GPU compute capability increases. This is because, unlike dense models, MoE models involve many memory-intensive operations like routing, local scatter, and gather, which remain time-consuming since memory bandwidth does not scale as quickly as compute capabilities. Additionally, GEMM efficiency declines with increasing compute capability, as it also relies on memory loading, constrained by memory bandwidth.

\subsection{Ablation Study}
\label{sec:evaluation:ablation}

\begin{table}[t!]
    \centering
    \resizebox{\linewidth}{!} {
        \begin{tabular}{c|c|c|c}
            \hline
            \hline
            \multirow{2}{*}{\begin{tabular}[c]{@{}c@{}}Idx\end{tabular}} & \multirow{2}{*}{\begin{tabular}[c]{@{}c@{}}Method\end{tabular}} & \multirow{2}{*}{\begin{tabular}[c]{@{}c@{}}Normalized\\ Throughput\end{tabular}} & \multirow{2}{*}{\begin{tabular}[c]{@{}c@{}} $\Delta$ \end{tabular}} \\
            & & & \\\cline{1-4}
            1 & baseline & 1 & \\
            2 & (1) with SP+EP & 1.13 & +13\% \\
            3 & (2) with inter-operator overlap & 1.22 & +9\% \\
            4 & (3) with intra-operator overlap & 1.28 & +6\% \\
            \hline
            \hline
        \end{tabular}
    }
    % \vspace{-0.1in}
    \caption{\revision{Throughput improvement breakdown when training the 352B MoE model with 240 NVIDIA H800 GPUs and batch size is 720.}}
    \vspace{-0.2in}
    \label{tab:evaluation:systematic_ablation}
\end{table}

\revision{We evaluate the effectiveness of the optimization techniques of \sysname. First, we conduct an experiment about systematic breakdown by incrementally enabling each technique to isolate its contribution to the overall performance. Table~\ref{tab:evaluation:systematic_ablation} shows the throughput improvement breakdown with different optimizations when
training the 352B MoE model on 240 GPUs with a global batch size of 720. The baseline is a version of \sysname that adopts TP for both attention and FFNs and disables communication-computation overlap. First, by applying communication-efficient strategies—namely, SP for attention and EP for experts—we achieve an initial 13\% throughput improvement over this baseline. We then target the primary bottleneck in large-scale MoE training: communication overhead. Our inter-operator and intra-operator overlap methods effectively hide these costs, further accelerating training by an additional 9\% and 6\%, respectively.}

\begin{figure}[t!]
    \centering
    \includegraphics[width=0.8\linewidth]{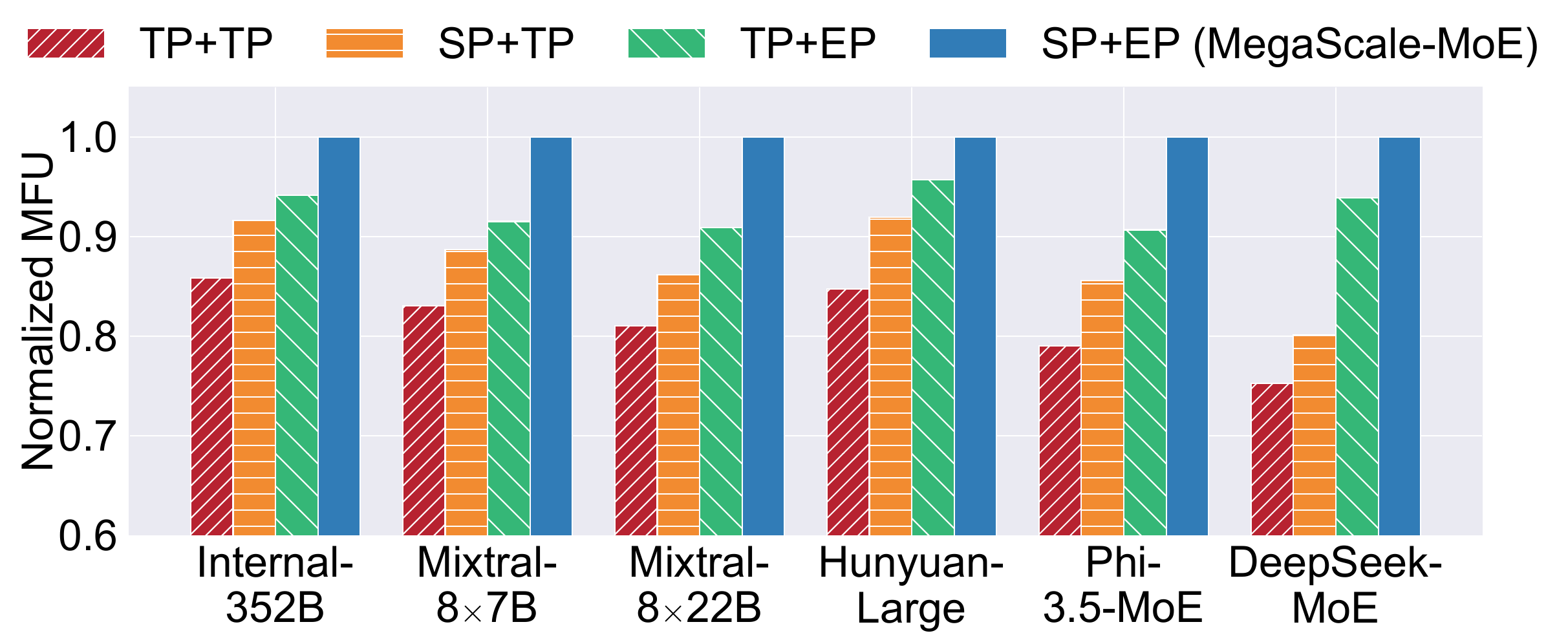}
    \vspace{-0.15in}
    \caption{Parallelism efficiency for different models.}
    \vspace{-0.1in}
    \label{fig:eval:parallelism}
\end{figure}

\begin{figure}[t!]
    \centering
    \includegraphics[width=0.9\linewidth]{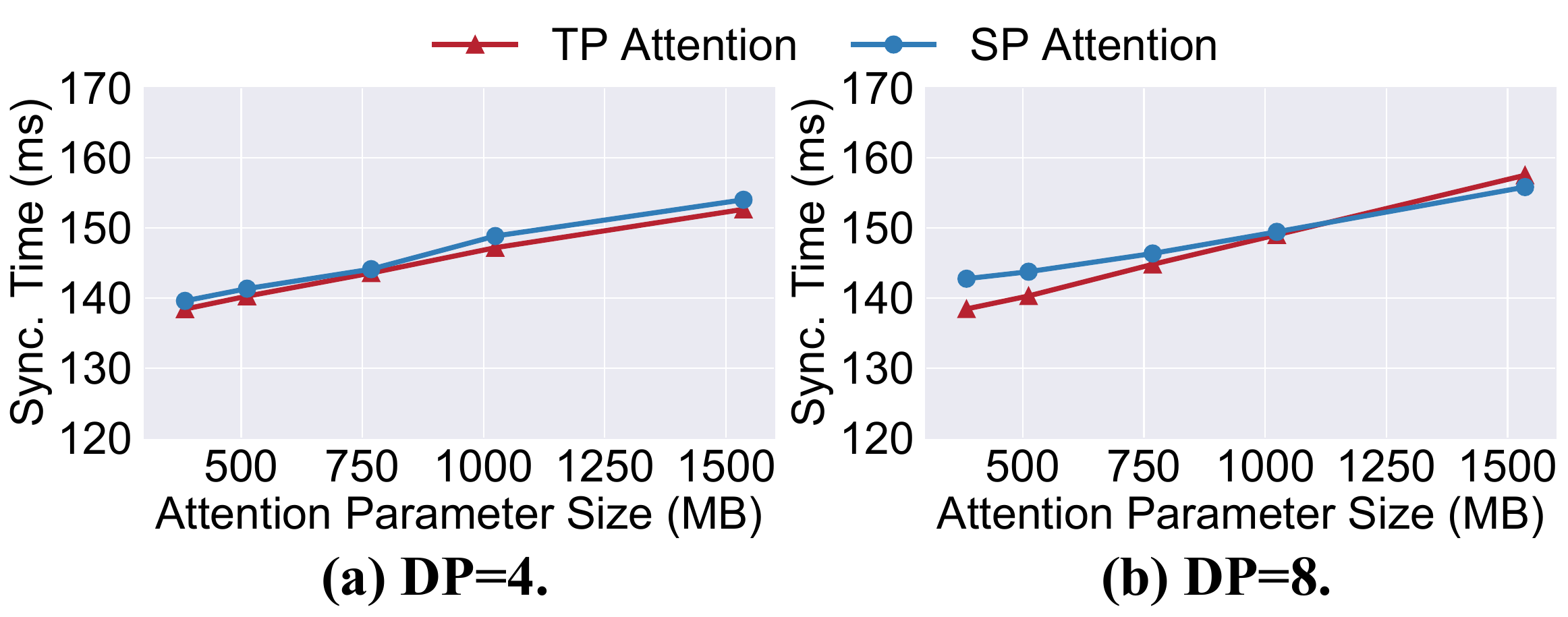}
    \vspace{-0.2in}
    \caption{Parameter synchronization time under SP and TP attention.}
    \vspace{-0.1in}
    \label{fig:eval:sync_latency}
\end{figure}

\revision{Following the systematic breakdown, we perform ablation studies on each component, varying a single setting at a time while keeping all others constant, to gain deeper insights into its behavior. }

\begin{figure*}[t!]
    \centering
    \includegraphics[width=\linewidth]{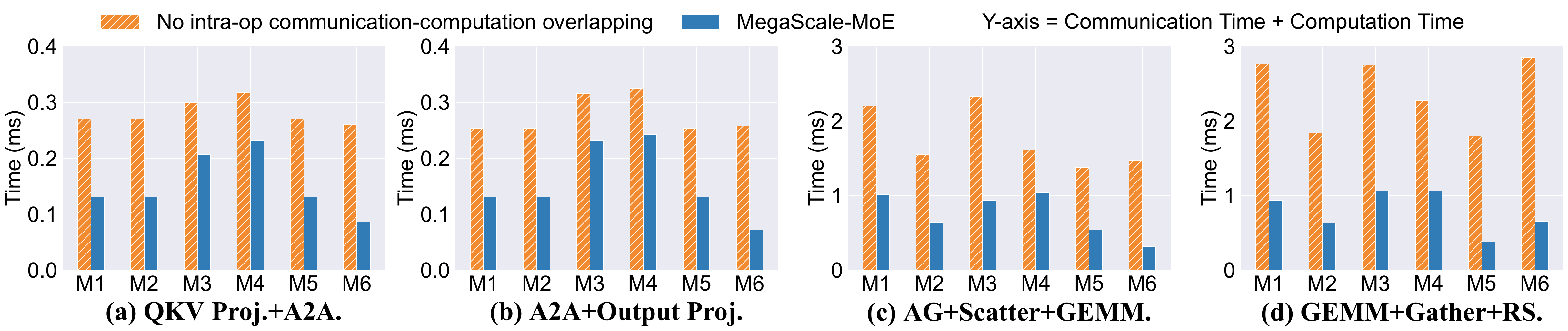}
    \vspace{-0.3in}
    \caption{Overlapped communication-computation time vs. non-overlapped time of each layer. M1-M6 represent the six models listed from top to bottom in Table~\ref{tab:evaluation:model_conf}; A2A, AG, and RS refer to all-to-all, all-gather, and reduce-scatter, respectively.}
    \vspace{-0.1in}
    \label{fig:eval:overlapping}
\end{figure*}

\parabf{Parallelism strategy.} We compare the training efficiency under various intra-node parallelism strategies using a single node with eight NVIDIA H800-SXM GPUs. We denote parallelism strategies as X+Y, where X represents the parallelism strategy for attention, and Y corresponds to that for experts. The available parallelism strategies for attention include TP and our SP, whereas for experts, the choices are TP and EP. To isolate the performance benefits of optimized parallelism, we disable other system optimizations. 

\begin{figure}[t!]
    \centering
    \includegraphics[width=\linewidth]{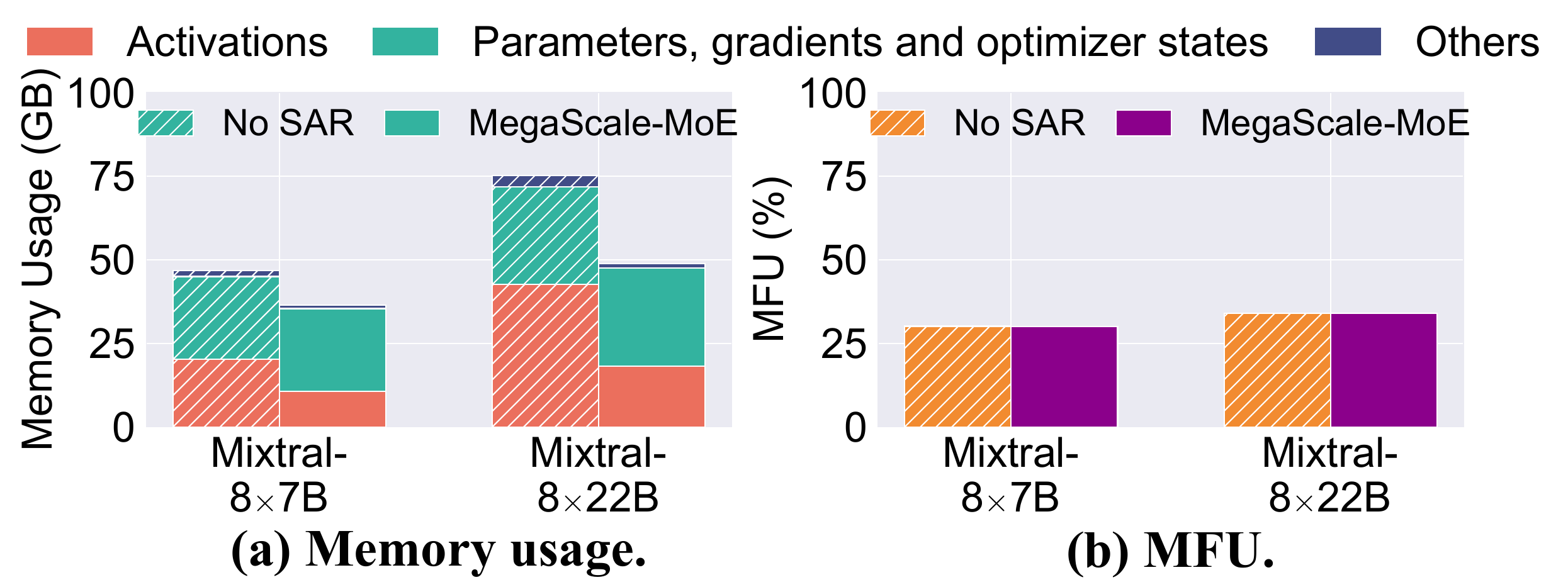}
    \vspace{-0.3in}
    \caption{Ablation study of selective activation rematerialization (SAR).}
    \vspace{-0.1in}
    \label{fig:eval:recompute}
\end{figure}

We measure the training MFU of one internal and five open-source MoE models with diverse model configurations as listed in Table~\ref{tab:evaluation:model_conf}. The global batch size is set to 32, and we adjust the number of layers for each model to fit within the GPU memory. Figure~\ref{fig:eval:parallelism} shows that \sysname's parallelism strategy, SP+EP, consistently outperforms the other three parallelism strategies, achieving 14.9\%-32.9\% higher MFU compared to TP+TP. The performance gains are attributed to two main factors. 
First, as discussed in \S\ref{sec:design:parallelism}, SP and EP effectively reduce the communication volume compared to TP, thereby decreasing communication overhead. Second, TP partitions the FFN module along the intermediate size dimension, which results in lower GEMM efficiency. 

To provide a more comprehensive evaluation of the parallelism strategy, we also report the additional overhead introduced by the replicated attention parameters in SP. In terms of memory usage, SP incurs a 1.2\%--5.4\% higher memory footprint compared to TP, requiring 1.7\%--8.1\% more memory to store parameters, gradients, and optimizer states across all seven models. This overhead is manageable considering the significant performance gains achieved by SP. 
% Besides, in large-scale training with pipeline parallelism, activation memory constitutes a substantial portion of the overall memory footprint, which reduces the impact of replicated attention parameters.

For the parameter synchronization time, we follow large-scale training setups and set the size of the TP or SP to 8, effectively parallelizing each layer within a single node. 
The attention parameter size on each GPU is varied from 384 MB to 1536 MB, while the FFN parameter size is fixed at 10 GB per GPU, reflecting typical real-world training setups. 
We run \sysname with SP and TP attention, using 4 and 8 DP groups, which correspond to a total of 32 and 64 GPUs, respectively.
Figure~\ref{fig:eval:sync_latency} shows that the synchronization times for SP and TP attention are consistently comparable, differing by only 0.3\%--3.1\%. This aligns with our hypothesis that SP and TP would exhibit similar performance characteristics in DP communication latency.

\parabf{Intra-operator commmunication overlap.} We then measure the duration of four key communication and the corresponding computation operators in the forward pass: $(i)$ QKV Projection paired with all-to-all, $(ii)$ all-to-all with Output Projection, $(iii)$ all-gather with scatter and GroupedGEMM, and $(iv)$ GroupedGEMM with gather and reduce-scatter, as depicted in Figure~\ref{fig:moe_computation}. Figure~\ref{fig:eval:overlapping} demonstrates that across all six models, \sysname achieves a 1.2–4.7$\times$ reduction in the combined time of communication and computation operators compared to the baseline lacking fine-grained overlap. And \sysname reduces the training iteration time by 7.1\%-12.9\% due to intra-operator communication-computation overlap.

\begin{figure}[t!]
    \centering
    \includegraphics[width=0.8\linewidth]{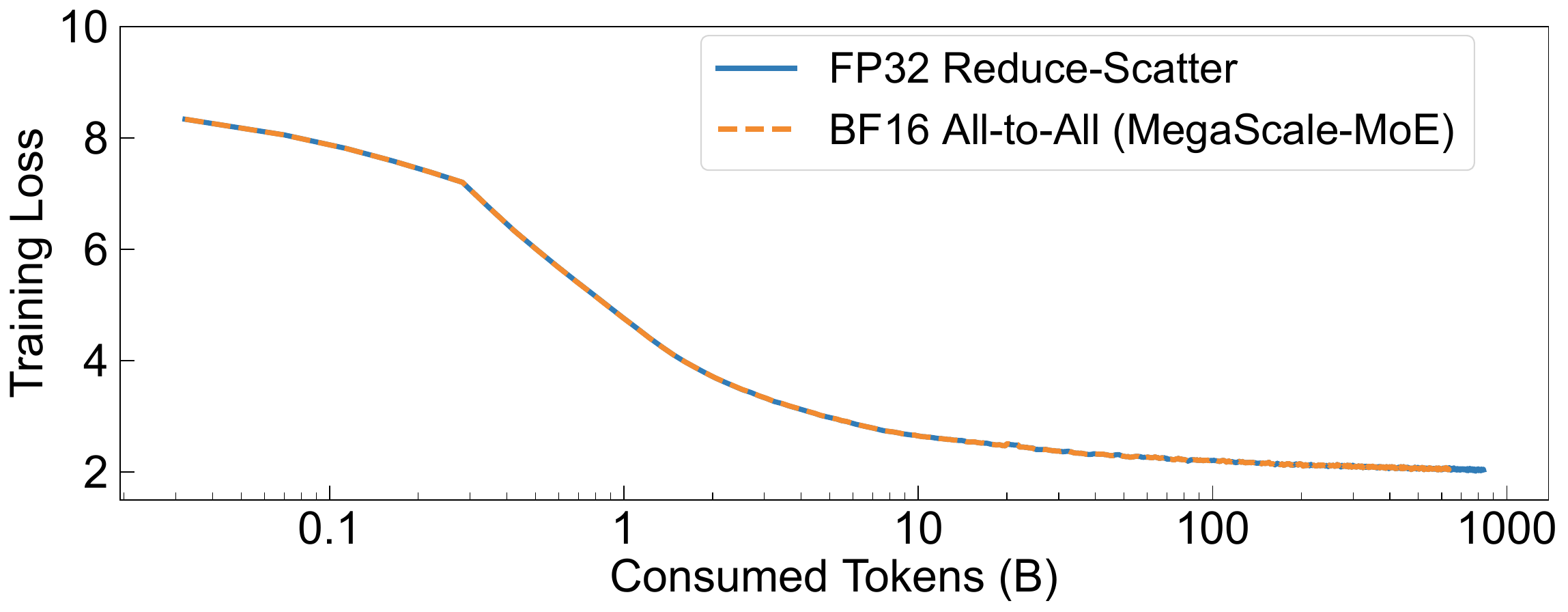}
    \vspace{-0.1in}
    \caption{The training loss curve of \sysname with DP communication compression.}
    \vspace{-0.1in}
    \label{fig:eval:bf16_a2a}
\end{figure}

\parabf{Selective activation rematerailization.} We compare \sysname to a baseline that disables selective activation rematerialization (No SAR), which stores all activations in GPU memory during training. We evaluate both methods by training Mixtral-8$\times$7B and Mixtral-8$\times$22B on 128 NVIDIA H800 GPUs. Figure~\ref{fig:eval:recompute} shows the memory usage breakdown and the training MFU. Compared to No SAR, \sysname reduces activation memory consumption by 45.5\% and 57.2\% for the two models, respectively, resulting in overall memory reductions of 21.3\% and 35\%, while maintaining the training performance difference within 0.5\%.

\parabf{Data parallelism communication compression.} We validate the effectiveness of our communication compression technique by training a 7B MoE model using BF16 all-to-all DP communication and FP32 reduce-scatter communication, as described in \S\ref{sec:design:precision}. Figure~\ref{fig:eval:bf16_a2a} illustrates the training loss curves, which are nearly identical. This optimization compresses only the accumulated gradients of the batch and performs conversions between BF16 and FP32 exclusively during communication, introducing minimal risk. 

\subsection{Model Convergence}
\label{sec:evaluation:co-design}

% We evaluate model convergence with our communication compression techniques. Figure~\ref{fig:eval:bf16_a2a} compares the training loss curves of \sysname using BF16 all-to-all DP communication and FP32 reduce-scatter communication. The two curves are nearly identical, as this optimization only compresses the gradients of the final micro-batch, posing minimal risk.
% We also validate the effectiveness of FP8 communication compression by training a 25B MoE model from scratch and continuing training a 100B MoE model from a checkpoint, with their loss curves shown in Figure~\ref{fig:eval:fp8}. With communication compression and optimizations such as tailored activation quantization, \sysname maintains convergence stability and achieves training loss consistent with the BF16 baseline.

\begin{figure}[t!]
    \centering
    \includegraphics[width=\linewidth]{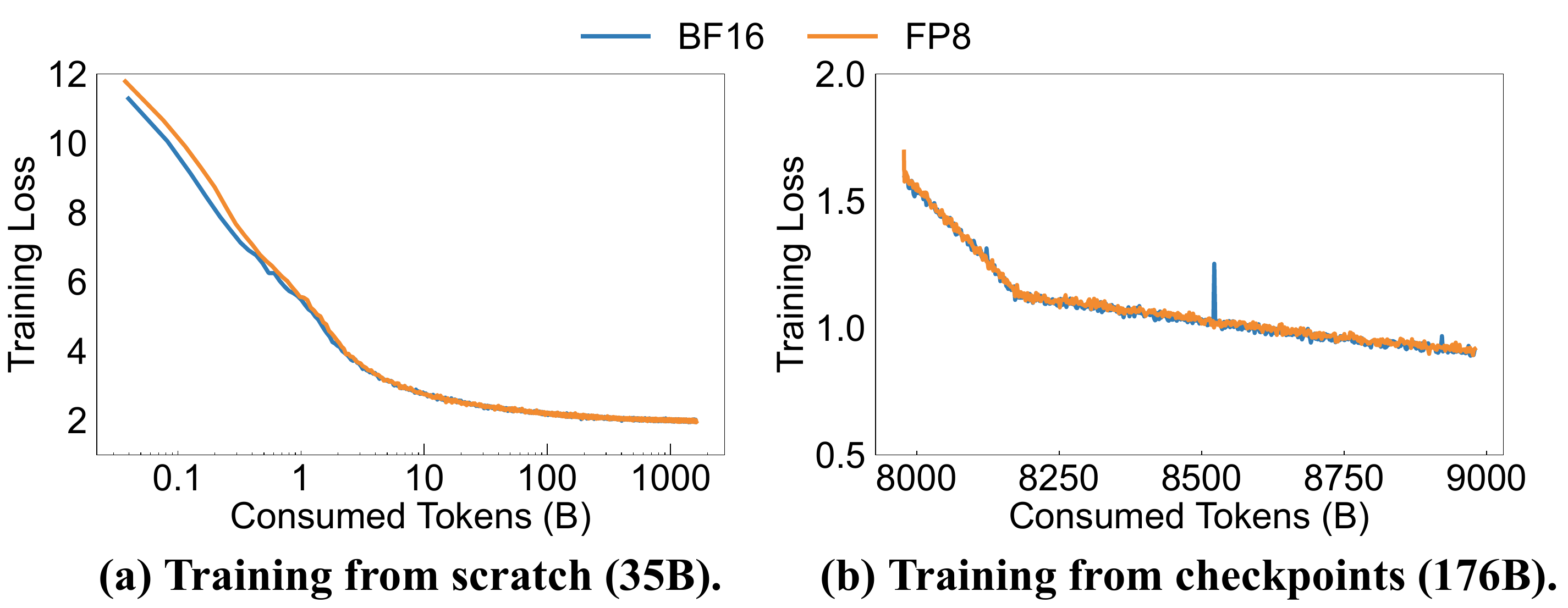}
    \vspace{-0.2in}
    \caption{The loss curve of \sysname in FP8 and BF16.}
    \vspace{-0.1in}
    \label{fig:eval:fp8}
\end{figure}

We evaluate model convergence with \sysname. Figure~\ref{fig:eval:fp8} demonstrates the loss curves of training a 35B MoE model from scratch and continuing training a 176B MoE model from a checkpoint, with results shown for both BF16 and FP8 precision. \sysname ensures stable convergence and consistent training loss across BF16 and FP8 formats.

\section{Experience}
\label{sec:experience}

In this section, we describe our deployment and operational experience of \sysname. 

\begin{figure}[t!]
    \centering
    \includegraphics[width=\linewidth]{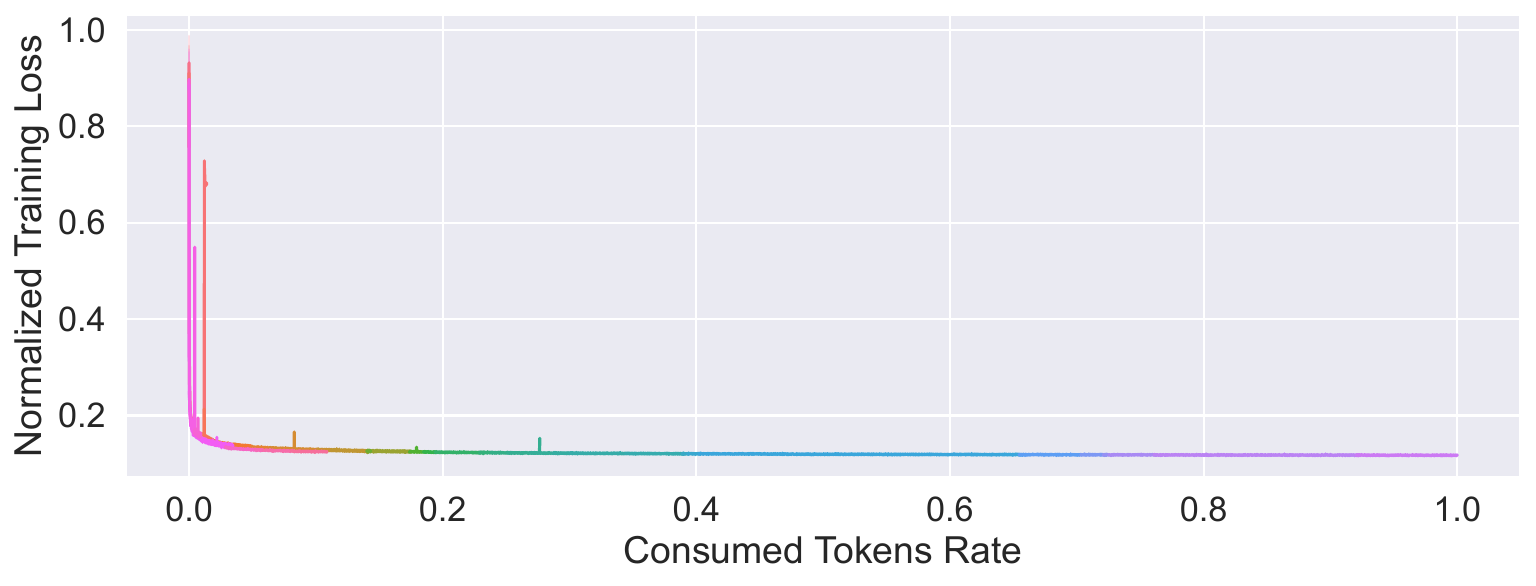}
    \vspace{-0.2in}
    \caption{The normalized training loss curve of a real production job on more than 10,000 GPUs for months, training a MoE model with 20B activated and 200B total parameters on multi-trillion tokens. Different colors indicate training restarts.}
    \vspace{-0.1in}
    \label{fig:experience:llm-loss}
\end{figure}

\parabf{Deployment experience.} \sysname has been deployed in our production environment and is responsible for the majority of large-scale MoE training tasks within our company. It enables the training of models with trillions of parameters, supports single training jobs scaling beyond 10,000 GPUs, with individual training tasks running for several months. By combining the aforementioned techniques, \sysname minimizes idle communication time and optimizes memory usage in MoE training without compromising model performance, ultimately saving millions of GPU hours in large-scale MoE training. 
Figure~\ref{fig:experience:llm-loss} shows the model convergence from a real production job, which trains a proprietary MoE model with 200B parameters, 20B activated for each token. This job uses over 10,000 GPUs and lasts for months. The loss continues to converge with a stable training process.

\parabf{FP8 training.} We have made extensive efforts to maintain the convergence stability of FP8 training. For example, we observe that the SwiGLU operator significantly expands the numerical range. To address this, we replace per-tensor quantization with higher-precision per-token quantization ($1 \times h$). Additionally, since multiplying SwiGLU with the gating weight further amplifies the dynamic numerical range, we shift the gating weight multiplication back to after the FC2 output, reducing quantization errors.

Beyond ensuring training convergence, we introduce additional engineering optimizations. Existing FP8 training implementations~\cite{transformer-engine, liang2024torchtitan} store model parameters in BF16, requiring frequent FP8 conversion for GEMM computations, adding casting and transpose overhead. To address this, we use a multi-precision optimizer to store model parameters directly in FP8, while keeping main parameters in FP32 with separate buffers for different data types. This lowers memory consumption and halves parameter all-gather communication in data parallelism.

\parabf{Scale up.} When training MoE models, an intriguing engineering question arises: can we indefinitely scale the training size by increasing model parameters without raising computational load? This approach is impractical in tensor parallelism, as scaling up the model necessitates a higher TP degree to accommodate additional parameters. While increased TP reduces per-GPU computation, the communication overhead remains constant, as shown in Formula~\ref{eq:1} and \ref{eq:4}, leading to progressively longer communication times and reduced training efficiency. In other words, TP has inherent scalability limitations and often relies on high-speed intra-node links to mitigate communication delays. 

In contrast, when scaling training with SP and EP, the communication volume decreases as the parallel size $n$ increases, as shown in Formula~\ref{eq:2} and \ref{eq:3}. This implies that, in theory, this parallelism strategy can scale to significantly larger sizes. However, in practical hierarchical infrastructures, a critical challenge emerges: can this approach maintain training efficiency when scaling beyond the NVLink domain, where bandwidth drops to RDMA levels?

Formally, for a SwiGLU structure incorporating a MoE mechanism, the ratio $R$ between computation time and communication time is defined as:
\begin{align}
    \text{comm\_time} & = \frac{2k \times bsh(n-1)/n /n}{bandwidth}, \\
    \text{comp\_time} & = \frac{3k \times bsh \times h_{ffn}/n}{peak}.
\end{align}
\begin{align}
    \hspace{-3cm} R &= \frac{\text{comp\_time}}{\text{comm\_time}} \\
    &= 3/2\times h_{ffn}\times \frac{bandwidth}{peak} \times n/(n-1) \\
    & \approx  3/2 \times h_{ffn} \times \frac{bandwidth}{peak}
\end{align}
To sustain training efficiency, the FFN's computation time must exceed the communication time, ensuring effective overlap of communication overhead. Therefore, our goal is to maintain $R > 1$, leading to two key insights:
\begin{itemize}[leftmargin=*]
    \item The value of $R$ is independent of the number of experts, top-$k$, hidden dimension, parallelism size, or input size, providing flexibility in selecting algorithm parameters.
    \item $R$ is solely determined by the expert's intermediate dimension, computational peak, and communication bandwidth. Consequently, on fixed hardware, as long as the expert dimension is sufficiently large, the MoE model can be scaled while maintaining training efficiency from an engineering perspective.
\end{itemize}

\parabf{Holistic vs. automatic.} We have invested substantial engineering efforts in inter-operator communication-computation overlap, including determining operator execution order, concurrency of communication and computation, and SM allocation for communication. These manual interventions provide deeper insights into training dynamics, enabling targeted optimizations. As training progresses and experience accumulates, we seek to automate operator scheduling within the search space to optimize the training process at a fine-grained level and achieve optimal performance. We leave automatic optimization for future work. 

\parabf{MoE vs. dense model training.} \revision{In our continued efforts to optimize MoE model training, we have identified several critical distinctions from the training of dense models. In a dense Transformer layer, optimization efforts are concentrated on self-attention and GEMMs. The former is often accelerated by techniques like FlashAttention~\cite{dao2022flashattention}, while the latter, as a dense computation, generally achieves high utilization on the GPU's parallel processing units. In contrast, as shown in Figure~\ref{fig:eval:different_gpus}a, the combined runtime of attention and GroupedGEMM accounts for only about one-third of a layer's execution time. The remainder is consumed by communication and other operators. While \sysname effectively addresses the communication overhead, we observe that the computational operators in MoE models, which are inherently more complex than their dense counterparts, also introduce performance degradation. Specifically, they are a primary source of stragglers for three main reasons:}

\revision{First, the intermediate dimension of each expert is smaller than the FFN layer in a dense model. To efficiently process computations for multiple experts concurrently, GroupedGEMM employs a single CUDA kernel for numerous small matrix multiplications. The resource usage of this kernel—including shared memory, L1 cache, and number of threads—is finely controlled via \texttt{cuFuncSetAttribute}. This granular control, however, can introduce synchronization delays. Second, due to the imbalanced number of tokens routed to each expert, the inputs and outputs for GroupedGEMM are dynamically shaped tensors. The frequent allocation and deallocation of these tensors exacerbate GPU memory fragmentation. Third, the MoE gating mechanism involves a multitude of small operators for tasks like calculating routing scores and communicating routing decisions. Jitter in CPU performance can delay the launch of these kernels to the point where the launch latency exceeds their actual execution time on the GPU, creating pipeline bubbles.}

% \subsection{stability, trouble-shooting}

% \section{Candidate Topics Below}

% \subsection{Performance Analyze Platform}

% \subsection{Straggler Detection}

% \subsection{Fast Recovery}

% \subsection{Low precesion training}
\section{Related Work}
\label{sec:related}

\paraf{Large model training.} LLM research has led to the development of scalable, efficient, and robust training techniques~\cite{rasley2020deepspeed, shoeybi2019megatron, jiang2024megascale, zhang2025disttrain} to meet the substantial computational demands of these models. DeepSpeed~\cite{rasley2020deepspeed} features the Zero Redundancy Optimizer (ZeRO)~\cite{rajbhandari2020zero, rajbhandari2021zero, ren2021zero}, which shards model parameters, gradients, and optimizer states across participating GPUs in data parallelism, enabling the scaling of LLMs with manageable memory consumption. Megatron-LM~\cite{shoeybi2019megatron} focuses on intra-layer model parallelism techniques, partitioning the parameters and computation of each layer. Pipeline parallelism assigns the parameters and computation of a contiguous subset of layers to each GPU\cite{huang2019gpipe, narayanan2019pipedream}, breaks a batch into micro-batches, and processes the micro-batches in a pipelined fashion. MegaScale~\cite{jiang2024megascale} shows how combining tensor, pipeline, and data parallelism can be an efficient strategy to train large multi-billion parameter models at unprecedented scale.

\parabf{Mixture-of-Expert training.} To address the computational challenges of training advanced neural networks, the machine learning field has increasingly adopted Mixture-of-Experts architectures. Subsequently, a number of deep learning frameworks have been proposed for training or running inference on MoEs on multi-GPU clusters. DeepSpeed-MoE~\cite{rajbhandari2022deepspeed} significantly reduces training costs through model architecture designs and compression techniques. HetuMoE~\cite{nie2022hetumoe} utilizes a hierarchical all-to-all communication strategy to achieve performance speedup. SE-MoE~\cite{shen2022se} distinguishes itself by focusing on scalable and efficient training with heterogeneous resources like CPU memory and SSDs.  FasterMoE~\cite{he2022fastermoe} introduces a comprehensive suite of optimizations such as dynamic shadowing, fine-grained scheduling, and congestion-avoiding expert selection strategies. Janus~\cite{liu2023janus} proposes a data-centric paradigm shift for MoE models, aiming to lower communication demands and boost training efficiency. \revision{Tutel~\cite{hwang2023tutel} offers a dynamic solution for MoE models, employing adaptive parallelism and pipelining.
However, its dynamic parallelism switching and hierarchical all-to-all can cause significant overheads for models with hundreds of billions of parameters. To avoid such overhead, latest MoE training systems~\cite{liu2024deepseek, liu2024deepseekv3} use auxiliary loss or routing bias for load balancing and limit cross-node token dispatch. By mapping each MoE layer to intra-node, \sysname eliminates cross-node token dispatch.}

\revision{Recently, DeepSeek-V3~\cite{liu2024deepseekv3} introduced two key optimizations for training production-scale MoE models: DeepEP, for high-performance cross-node all-to-all communication, and DualPipe, for overlapping communication with computation. Due to the relatively low cross-node InfiniBand bandwidth, DeepEP limits the token dispatch to a maximum of 4 nodes to maintain a constant cross-node communication volume, restricting its routing flexibility. In contrast, \sysname places each MoE layer intra-node to ensure efficient routing to any top-k experts. DualPipe leverages pipeline parallelism for communication-computation overlap across different micro-batches, which requires storing 2$\times$ the model parameters. In contrast, \sysname's overlap occurs within a single micro-batch's forward or backward pass, incurring no additional memory overhead and remaining compatible with systems both with and without pipeline parallelism.}

\parabf{Long-context training.} While Megatron-LM~\cite{shoeybi2019megatron, korthikanti2023reducing} opts to partition only specific operations along the sequence dimension, various methods of sequence parallelism~\cite{li2024distflashattn, liu2023ring, li2021sequence, gu2024loongtrain} have been explored for training models requiring long contexts. The Blockwise Parallel Transformer~\cite{liu2024blockwise} method implements blockwise computation of self-attention and the fusion of FFNs based on online softmax calculations. Ring Attention~\cite{liu2023ring, li2021sequence} introduces a ring-style communication mechanism integrated with self-attention calculations, facilitating the exchange of key and value chunks. 
% Compared to these methods, our Sequence Parallel Attention adopts several unique design choices to accommodate our target scenario: we distribute the FFN's weights with other parallelism given its significant memory footprint. We opt for a single all-gather operation to retain the full key-value pairs in memory, rather than re-communicating them during the backward pass. This strategy enables better performance in scenarios without extended context requirements.
We adopt the all-to-all style of SP attention from DeepSpeed Ulysses~\cite{jacobs2023deepspeed}, which partitions attention by heads rather than sequence length, due to its reduced communication volume and balanced computation pattern.

\parabf{Communication-computation overlap.} Several frameworks~\cite{hashemi2019tictac, li2020pytorch, mahajan2023better, peng2019generic, pytorch_fsdp} focus on overlapping communication with computation in distributed deep learning training with a single parallelism strategy. Some compiler-style work~\cite{jangda2022breaking, wang2022overlap, pati2024t3} provides fine-grained overlap among kernels, but excessive partitioning of GEMM kernels can result in low GPU utilization. Centauri~\cite{chen2024centauri} enhances communication overlap for LLM training with 3D parallelism by communication partitioning and hierarchical scheduling. Similar to Centauri, our inter-operator communication overlap hides communication within independent computation by reordering operators. We further conceal communication on critical paths through intra-operator overlap, without compromising GPU utilization. 

% DualPipe, proposed by DeepSeek-V3~\cite{liu2024deepseekv3}, overlaps communication and computation within different forward and backward chunks and requires storing 2$\times$ the model parameters. In contrast, \sysname achieves this overlap within a single forward or backward chunk without incurring additional memory overhead.
\section{Conclusion}
\label{sec:conclusion}

In this paper, we offer an in-depth look at the design, implementation, and deployment of \sysname, a production-grade system built to efficiently train MoE models. \sysname exploits communication-efficient approaches, including parallelism strategies with lower communication volume, inter- and intra-operator communication-computation overlap, and communication compression with adjusted communication patterns to unleash the compute capabilities of high-performance GPUs. \sysname achieves 1.41M tokens/s in throughput when training a 352B MoE model on 1,440 NVIDIA Hopper GPUs, a 1.88$\times$ improvement over Megatron-LM. By sharing our insights on accelerating large-scale MoE training, we hope our work will inspire future research.

% We emphasize the need for fault tolerance throughout the training process and implement a tailored robust training framework to locate and fix faults automatically. We provide a comprehensive set of monitoring tools for deep observability into system components and events, facilitating root cause identification for intricate anomalies. We believe that our work not only offers practical insights for those working on LLM training, but also paves the way for future research in this rapidly evolving field.

\parabf{Acknowledgements.}
We thank our shepherd, Cheng Li, and the anonymous reviewers for their valuable feedback and suggestions.
This work was supported in part by the National Key Research and Development Program of China under Grant 2022YFB4500700, the Scientific Research Innovation Capability Support Project for Young Faculty under Grant ZYGXQNJSKYCXNLZCXM-I1, the Fundamental Research Funds for the Central Universities, Peking University, and the National Natural Science Foundation of China under Grant 62172008 and Grant 62325201.
Xin Jin and Xin Liu are the corresponding authors.
Chao Jin, Xuanzhe Liu, and Xin Jin are also with the Key Laboratory of High Confidence Software Technologies (Peking University), Ministry of Education.

\label{lastpage}

\def\UrlBreaks{\do\/\do-}
\bibliographystyle{ACM-Reference-Format}
\bibliography{reference}


\begin{thebibliography}{56}

%%% the Web address.

\ifx \showCODEN    \undefined \def \showCODEN     #1{\unskip}     \fi
\ifx \showDOI      \undefined \def \showDOI       #1{#1}\fi
\ifx \showISBNx    \undefined \def \showISBNx     #1{\unskip}     \fi
\ifx \showISBNxiii \undefined \def \showISBNxiii  #1{\unskip}     \fi
\ifx \showISSN     \undefined \def \showISSN      #1{\unskip}     \fi
\ifx \showLCCN     \undefined \def \showLCCN      #1{\unskip}     \fi
\ifx \shownote     \undefined \def \shownote      #1{#1}          \fi
\ifx \showarticletitle \undefined \def \showarticletitle #1{#1}   \fi
\ifx \showURL      \undefined \def \showURL       {\relax}        \fi
% invisible to TeX
\providecommand\bibfield[2]{#2}
\providecommand\bibinfo[2]{#2}
\providecommand\natexlab[1]{#1}
\providecommand\showeprint[2][]{arXiv:#2}

\bibitem[\protect\citeauthoryear{??}{con}{2025}]%
        {context_parallelism}
 \bibinfo{year}{2025}\natexlab{}.
\newblock \bibinfo{title}{Context parallelism in Megatron-LM}.
\newblock   (\bibinfo{year}{2025}).
\newblock
\newblock
\shownote{\url{https://docs.nvidia.com/megatron-core/developer-guide/latest/api-guide/context_parallel.html}.}


\bibitem[\protect\citeauthoryear{??}{dbr}{2025}]%
        {dbrx}
 \bibinfo{year}{2025}\natexlab{}.
\newblock \bibinfo{title}{Introducing DBRX: A New State-of-the-Art Open LLM}.
\newblock   (\bibinfo{year}{2025}).
\newblock
\showURL{%
\url{https://www.databricks.com/blog/introducing-dbrx-new-state-art-open-llm}}


\bibitem[\protect\citeauthoryear{??}{gro}{2025}]%
        {grok}
 \bibinfo{year}{2025}\natexlab{}.
\newblock \bibinfo{title}{Open Release of Grok-1}.
\newblock   (\bibinfo{year}{2025}).
\newblock
\showURL{%
\url{https://x.ai/blog/grok-os}}


\bibitem[\protect\citeauthoryear{Ainslie, Lee-Thorp, de~Jong, Zemlyanskiy,
  Lebr{\'o}n, and Sanghai}{Ainslie et~al\mbox{.}}{2023}]%
        {ainslie2023gqa}
\bibfield{author}{\bibinfo{person}{Joshua Ainslie}, \bibinfo{person}{James
  Lee-Thorp}, \bibinfo{person}{Michiel de Jong}, \bibinfo{person}{Yury
  Zemlyanskiy}, \bibinfo{person}{Federico Lebr{\'o}n}, {and}
  \bibinfo{person}{Sumit Sanghai}.} \bibinfo{year}{2023}\natexlab{}.
\newblock \showarticletitle{Gqa: Training generalized multi-query transformer
  models from multi-head checkpoints}.
\newblock \bibinfo{journal}{{\em arXiv preprint arXiv:2305.13245\/}}
  (\bibinfo{year}{2023}).
\newblock


\bibitem[\protect\citeauthoryear{Chang, Bao, Hou, Jiang, Zheng, Zhong, Zhang,
  Song, Yao, Jiang, et~al\mbox{.}}{Chang et~al\mbox{.}}{2024}]%
        {chang2024flux}
\bibfield{author}{\bibinfo{person}{Li-Wen Chang}, \bibinfo{person}{Wenlei Bao},
  \bibinfo{person}{Qi Hou}, \bibinfo{person}{Chengquan Jiang},
  \bibinfo{person}{Ningxin Zheng}, \bibinfo{person}{Yinmin Zhong},
  \bibinfo{person}{Xuanrun Zhang}, \bibinfo{person}{Zuquan Song},
  \bibinfo{person}{Chengji Yao}, \bibinfo{person}{Ziheng Jiang},
  {et~al\mbox{.}}} \bibinfo{year}{2024}\natexlab{}.
\newblock \showarticletitle{FLUX: fast software-based communication overlap on
  gpus through kernel fusion}.
\newblock \bibinfo{journal}{{\em arXiv preprint arXiv:2406.06858\/}}
  (\bibinfo{year}{2024}).
\newblock


\bibitem[\protect\citeauthoryear{Chen, Li, Zhu, Duan, Sun, Zhang, and
  Yang}{Chen et~al\mbox{.}}{2024}]%
        {chen2024centauri}
\bibfield{author}{\bibinfo{person}{Chang Chen}, \bibinfo{person}{Xiuhong Li},
  \bibinfo{person}{Qianchao Zhu}, \bibinfo{person}{Jiangfei Duan},
  \bibinfo{person}{Peng Sun}, \bibinfo{person}{Xingcheng Zhang}, {and}
  \bibinfo{person}{Chao Yang}.} \bibinfo{year}{2024}\natexlab{}.
\newblock \showarticletitle{Centauri: Enabling Efficient Scheduling for
  Communication-Computation Overlap in Large Model Training via Communication
  Partitioning}. In \bibinfo{booktitle}{{\em ACM ASPLOS}}.
\newblock


\bibitem[\protect\citeauthoryear{Chowdhery, Narang, Devlin, Bosma, Mishra,
  Roberts, Barham, Chung, Sutton, Gehrmann, et~al\mbox{.}}{Chowdhery
  et~al\mbox{.}}{2023}]%
        {chowdhery2023palm}
\bibfield{author}{\bibinfo{person}{Aakanksha Chowdhery},
  \bibinfo{person}{Sharan Narang}, \bibinfo{person}{Jacob Devlin},
  \bibinfo{person}{Maarten Bosma}, \bibinfo{person}{Gaurav Mishra},
  \bibinfo{person}{Adam Roberts}, \bibinfo{person}{Paul Barham},
  \bibinfo{person}{Hyung~Won Chung}, \bibinfo{person}{Charles Sutton},
  \bibinfo{person}{Sebastian Gehrmann}, {et~al\mbox{.}}}
  \bibinfo{year}{2023}\natexlab{}.
\newblock \showarticletitle{Palm: Scaling language modeling with pathways}.
\newblock \bibinfo{journal}{{\em Journal of Machine Learning Research\/}}
  (\bibinfo{year}{2023}).
\newblock


\bibitem[\protect\citeauthoryear{Dao, Fu, Ermon, Rudra, and R{\'e}}{Dao
  et~al\mbox{.}}{2022}]%
        {dao2022flashattention}
\bibfield{author}{\bibinfo{person}{Tri Dao}, \bibinfo{person}{Dan Fu},
  \bibinfo{person}{Stefano Ermon}, \bibinfo{person}{Atri Rudra}, {and}
  \bibinfo{person}{Christopher R{\'e}}.} \bibinfo{year}{2022}\natexlab{}.
\newblock \showarticletitle{Flashattention: Fast and memory-efficient exact
  attention with io-awareness}.
\newblock \bibinfo{journal}{{\em Neural Information Processing Systems\/}}
  (\bibinfo{year}{2022}).
\newblock


\bibitem[\protect\citeauthoryear{Du, Huang, Dai, Tong, Lepikhin, Xu, Krikun,
  Zhou, Yu, Firat, Zoph, Fedus, Bosma, Zhou, Wang, Wang, Webster, Pellat,
  Robinson, Meier-Hellstern, Duke, Dixon, Zhang, Le, Wu, Chen, and Cui}{Du
  et~al\mbox{.}}{2022}]%
        {google_GLaM}
\bibfield{author}{\bibinfo{person}{Nan Du}, \bibinfo{person}{Yanping Huang},
  \bibinfo{person}{Andrew~M Dai}, \bibinfo{person}{Simon Tong},
  \bibinfo{person}{Dmitry Lepikhin}, \bibinfo{person}{Yuanzhong Xu},
  \bibinfo{person}{Maxim Krikun}, \bibinfo{person}{Yanqi Zhou},
  \bibinfo{person}{Adams~Wei Yu}, \bibinfo{person}{Orhan Firat},
  \bibinfo{person}{Barret Zoph}, \bibinfo{person}{Liam Fedus},
  \bibinfo{person}{Maarten~P Bosma}, \bibinfo{person}{Zongwei Zhou},
  \bibinfo{person}{Tao Wang}, \bibinfo{person}{Emma Wang},
  \bibinfo{person}{Kellie Webster}, \bibinfo{person}{Marie Pellat},
  \bibinfo{person}{Kevin Robinson}, \bibinfo{person}{Kathleen Meier-Hellstern},
  \bibinfo{person}{Toju Duke}, \bibinfo{person}{Lucas Dixon},
  \bibinfo{person}{Kun Zhang}, \bibinfo{person}{Quoc Le},
  \bibinfo{person}{Yonghui Wu}, \bibinfo{person}{Zhifeng Chen}, {and}
  \bibinfo{person}{Claire Cui}.} \bibinfo{year}{2022}\natexlab{}.
\newblock \showarticletitle{{GL}a{M}: Efficient Scaling of Language Models with
  Mixture-of-Experts}. In \bibinfo{booktitle}{{\em International Conference on
  Machine Learning (ICML)}}.
\newblock


\bibitem[\protect\citeauthoryear{Fedus, Zoph, and Shazeer}{Fedus
  et~al\mbox{.}}{2022}]%
        {fedus2022switch}
\bibfield{author}{\bibinfo{person}{William Fedus}, \bibinfo{person}{Barret
  Zoph}, {and} \bibinfo{person}{Noam Shazeer}.}
  \bibinfo{year}{2022}\natexlab{}.
\newblock \showarticletitle{Switch transformers: Scaling to trillion parameter
  models with simple and efficient sparsity}.
\newblock \bibinfo{journal}{{\em Journal of Machine Learning Research\/}}
  (\bibinfo{year}{2022}).
\newblock


\bibitem[\protect\citeauthoryear{Gu, Sun, Hu, Huang, Chen, Xiong, Wang, Chen,
  Zhao, Fang, et~al\mbox{.}}{Gu et~al\mbox{.}}{2024}]%
        {gu2024loongtrain}
\bibfield{author}{\bibinfo{person}{Diandian Gu}, \bibinfo{person}{Peng Sun},
  \bibinfo{person}{Qinghao Hu}, \bibinfo{person}{Ting Huang},
  \bibinfo{person}{Xun Chen}, \bibinfo{person}{Yingtong Xiong},
  \bibinfo{person}{Guoteng Wang}, \bibinfo{person}{Qiaoling Chen},
  \bibinfo{person}{Shangchun Zhao}, \bibinfo{person}{Jiarui Fang},
  {et~al\mbox{.}}} \bibinfo{year}{2024}\natexlab{}.
\newblock \showarticletitle{Loongtrain: Efficient training of long-sequence
  llms with head-context parallelism}.
\newblock \bibinfo{journal}{{\em arXiv preprint arXiv:2406.18485\/}}
  (\bibinfo{year}{2024}).
\newblock


\bibitem[\protect\citeauthoryear{Hashemi, Abdu~Jyothi, and Campbell}{Hashemi
  et~al\mbox{.}}{2019}]%
        {hashemi2019tictac}
\bibfield{author}{\bibinfo{person}{Sayed~Hadi Hashemi},
  \bibinfo{person}{Sangeetha Abdu~Jyothi}, {and} \bibinfo{person}{Roy
  Campbell}.} \bibinfo{year}{2019}\natexlab{}.
\newblock \showarticletitle{Tictac: Accelerating distributed deep learning with
  communication scheduling}.
\newblock \bibinfo{journal}{{\em Proceedings of Machine Learning and
  Systems\/}} (\bibinfo{year}{2019}).
\newblock


\bibitem[\protect\citeauthoryear{He, Zhai, Antunes, Wang, Luo, Shi, and Li}{He
  et~al\mbox{.}}{2022}]%
        {he2022fastermoe}
\bibfield{author}{\bibinfo{person}{Jiaao He}, \bibinfo{person}{Jidong Zhai},
  \bibinfo{person}{Tiago Antunes}, \bibinfo{person}{Haojie Wang},
  \bibinfo{person}{Fuwen Luo}, \bibinfo{person}{Shangfeng Shi}, {and}
  \bibinfo{person}{Qin Li}.} \bibinfo{year}{2022}\natexlab{}.
\newblock \showarticletitle{Fastermoe: modeling and optimizing training of
  large-scale dynamic pre-trained models}. In \bibinfo{booktitle}{{\em ACM
  PPoPP}}.
\newblock


\bibitem[\protect\citeauthoryear{Huang, Cheng, Bapna, Firat, Chen, Chen, Lee,
  Ngiam, Le, Wu, et~al\mbox{.}}{Huang et~al\mbox{.}}{2019}]%
        {huang2019gpipe}
\bibfield{author}{\bibinfo{person}{Yanping Huang}, \bibinfo{person}{Youlong
  Cheng}, \bibinfo{person}{Ankur Bapna}, \bibinfo{person}{Orhan Firat},
  \bibinfo{person}{Dehao Chen}, \bibinfo{person}{Mia Chen},
  \bibinfo{person}{HyoukJoong Lee}, \bibinfo{person}{Jiquan Ngiam},
  \bibinfo{person}{Quoc~V Le}, \bibinfo{person}{Yonghui Wu}, {et~al\mbox{.}}}
  \bibinfo{year}{2019}\natexlab{}.
\newblock \showarticletitle{Gpipe: Efficient training of giant neural networks
  using pipeline parallelism}.
\newblock \bibinfo{journal}{{\em Neural Information Processing Systems\/}}
  (\bibinfo{year}{2019}).
\newblock


\bibitem[\protect\citeauthoryear{Hwang, Cui, Xiong, Yang, Liu, Hu, Wang, Salas,
  Jose, Ram, et~al\mbox{.}}{Hwang et~al\mbox{.}}{2023}]%
        {hwang2023tutel}
\bibfield{author}{\bibinfo{person}{Changho Hwang}, \bibinfo{person}{Wei Cui},
  \bibinfo{person}{Yifan Xiong}, \bibinfo{person}{Ziyue Yang},
  \bibinfo{person}{Ze Liu}, \bibinfo{person}{Han Hu}, \bibinfo{person}{Zilong
  Wang}, \bibinfo{person}{Rafael Salas}, \bibinfo{person}{Jithin Jose},
  \bibinfo{person}{Prabhat Ram}, {et~al\mbox{.}}}
  \bibinfo{year}{2023}\natexlab{}.
\newblock \showarticletitle{Tutel: Adaptive mixture-of-experts at scale}.
\newblock \bibinfo{journal}{{\em Proceedings of Machine Learning and
  Systems\/}} (\bibinfo{year}{2023}).
\newblock


\bibitem[\protect\citeauthoryear{Jacobs, Tanaka, Zhang, Zhang, Song,
  Rajbhandari, and He}{Jacobs et~al\mbox{.}}{2023}]%
        {jacobs2023deepspeed}
\bibfield{author}{\bibinfo{person}{Sam~Ade Jacobs}, \bibinfo{person}{Masahiro
  Tanaka}, \bibinfo{person}{Chengming Zhang}, \bibinfo{person}{Minjia Zhang},
  \bibinfo{person}{Shuaiwen~Leon Song}, \bibinfo{person}{Samyam Rajbhandari},
  {and} \bibinfo{person}{Yuxiong He}.} \bibinfo{year}{2023}\natexlab{}.
\newblock \showarticletitle{Deepspeed ulysses: System optimizations for
  enabling training of extreme long sequence transformer models}.
\newblock \bibinfo{journal}{{\em arXiv preprint arXiv:2309.14509\/}}
  (\bibinfo{year}{2023}).
\newblock


\bibitem[\protect\citeauthoryear{Jangda, Huang, Liu, Sabet, Maleki, Miao,
  Musuvathi, Mytkowicz, and Saarikivi}{Jangda et~al\mbox{.}}{2022}]%
        {jangda2022breaking}
\bibfield{author}{\bibinfo{person}{Abhinav Jangda}, \bibinfo{person}{Jun
  Huang}, \bibinfo{person}{Guodong Liu}, \bibinfo{person}{Amir Hossein~Nodehi
  Sabet}, \bibinfo{person}{Saeed Maleki}, \bibinfo{person}{Youshan Miao},
  \bibinfo{person}{Madanlal Musuvathi}, \bibinfo{person}{Todd Mytkowicz}, {and}
  \bibinfo{person}{Olli Saarikivi}.} \bibinfo{year}{2022}\natexlab{}.
\newblock \showarticletitle{Breaking the computation and communication
  abstraction barrier in distributed machine learning workloads}. In
  \bibinfo{booktitle}{{\em ACM ASPLOS}}.
\newblock


\bibitem[\protect\citeauthoryear{Jiang, Sablayrolles, Roux, Mensch, Savary,
  Bamford, Chaplot, Casas, Hanna, Bressand, et~al\mbox{.}}{Jiang
  et~al\mbox{.}}{2024b}]%
        {jiang2024mixtral}
\bibfield{author}{\bibinfo{person}{Albert~Q Jiang}, \bibinfo{person}{Alexandre
  Sablayrolles}, \bibinfo{person}{Antoine Roux}, \bibinfo{person}{Arthur
  Mensch}, \bibinfo{person}{Blanche Savary}, \bibinfo{person}{Chris Bamford},
  \bibinfo{person}{Devendra~Singh Chaplot}, \bibinfo{person}{Diego de~las
  Casas}, \bibinfo{person}{Emma~Bou Hanna}, \bibinfo{person}{Florian Bressand},
  {et~al\mbox{.}}} \bibinfo{year}{2024}\natexlab{b}.
\newblock \showarticletitle{Mixtral of experts}.
\newblock \bibinfo{journal}{{\em arXiv preprint arXiv:2401.04088\/}}
  (\bibinfo{year}{2024}).
\newblock


\bibitem[\protect\citeauthoryear{Jiang, Lin, Zhong, Huang, Chen, Zhang, Peng,
  Li, Xie, Nong, Jia, He, Chen, Bai, Hou, Yan, Zhou, Sheng, Jiang, Xu, Wei,
  Zhang, Nie, Zou, Zhao, Xiang, Liu, Li, Jia, Ye, Jin, and Liu}{Jiang
  et~al\mbox{.}}{2024a}]%
        {jiang2024megascale}
\bibfield{author}{\bibinfo{person}{Ziheng Jiang}, \bibinfo{person}{Haibin Lin},
  \bibinfo{person}{Yinmin Zhong}, \bibinfo{person}{Qi Huang},
  \bibinfo{person}{Yangrui Chen}, \bibinfo{person}{Zhi Zhang},
  \bibinfo{person}{Yanghua Peng}, \bibinfo{person}{Xiang Li},
  \bibinfo{person}{Cong Xie}, \bibinfo{person}{Shibiao Nong},
  \bibinfo{person}{Yulu Jia}, \bibinfo{person}{Sun He},
  \bibinfo{person}{Hongmin Chen}, \bibinfo{person}{Zhihao Bai},
  \bibinfo{person}{Qi Hou}, \bibinfo{person}{Shipeng Yan},
  \bibinfo{person}{Ding Zhou}, \bibinfo{person}{Yiyao Sheng},
  \bibinfo{person}{Zhuo Jiang}, \bibinfo{person}{Haohan Xu},
  \bibinfo{person}{Haoran Wei}, \bibinfo{person}{Zhang Zhang},
  \bibinfo{person}{Pengfei Nie}, \bibinfo{person}{Leqi Zou},
  \bibinfo{person}{Sida Zhao}, \bibinfo{person}{Liang Xiang},
  \bibinfo{person}{Zherui Liu}, \bibinfo{person}{Zhe Li},
  \bibinfo{person}{Xiaoying Jia}, \bibinfo{person}{Jianxi Ye},
  \bibinfo{person}{Xin Jin}, {and} \bibinfo{person}{Xin Liu}.}
  \bibinfo{year}{2024}\natexlab{a}.
\newblock \showarticletitle{{MegaScale}: Scaling Large Language Model Training
  to More Than 10,000 {GPUs}}. In \bibinfo{booktitle}{{\em USENIX NSDI}}.
\newblock


\bibitem[\protect\citeauthoryear{Korthikanti, Casper, Lym, McAfee, Andersch,
  Shoeybi, and Catanzaro}{Korthikanti et~al\mbox{.}}{2023}]%
        {korthikanti2023reducing}
\bibfield{author}{\bibinfo{person}{Vijay~Anand Korthikanti},
  \bibinfo{person}{Jared Casper}, \bibinfo{person}{Sangkug Lym},
  \bibinfo{person}{Lawrence McAfee}, \bibinfo{person}{Michael Andersch},
  \bibinfo{person}{Mohammad Shoeybi}, {and} \bibinfo{person}{Bryan Catanzaro}.}
  \bibinfo{year}{2023}\natexlab{}.
\newblock \showarticletitle{Reducing activation recomputation in large
  transformer models}.
\newblock \bibinfo{journal}{{\em Proceedings of Machine Learning and
  Systems\/}} (\bibinfo{year}{2023}).
\newblock


\bibitem[\protect\citeauthoryear{Li, Shao, Xie, Xing, Ma, Stoica, Gonzalez, and
  Zhang}{Li et~al\mbox{.}}{2024}]%
        {li2024distflashattn}
\bibfield{author}{\bibinfo{person}{Dacheng Li}, \bibinfo{person}{Rulin Shao},
  \bibinfo{person}{Anze Xie}, \bibinfo{person}{Eric~P. Xing},
  \bibinfo{person}{Xuezhe Ma}, \bibinfo{person}{Ion Stoica},
  \bibinfo{person}{Joseph~E. Gonzalez}, {and} \bibinfo{person}{Hao Zhang}.}
  \bibinfo{year}{2024}\natexlab{}.
\newblock \showarticletitle{DISTFLASHATTN: Distributed Memory-efficient
  Attention for Long-context LLMs Training}.
\newblock \bibinfo{journal}{{\em arxiv preprint arXiv:2310.03294\/}}
  (\bibinfo{year}{2024}).
\newblock


\bibitem[\protect\citeauthoryear{Li, Jiang, Zhu, Wang, and Xu}{Li
  et~al\mbox{.}}{2023}]%
        {li2023accelerating}
\bibfield{author}{\bibinfo{person}{Jiamin Li}, \bibinfo{person}{Yimin Jiang},
  \bibinfo{person}{Yibo Zhu}, \bibinfo{person}{Cong Wang}, {and}
  \bibinfo{person}{Hong Xu}.} \bibinfo{year}{2023}\natexlab{}.
\newblock \showarticletitle{Accelerating distributed $\{$MoE$\}$ training and
  inference with lina}. In \bibinfo{booktitle}{{\em USENIX ATC}}.
\newblock


\bibitem[\protect\citeauthoryear{Li, Xue, Baranwal, Li, and You}{Li
  et~al\mbox{.}}{2021}]%
        {li2021sequence}
\bibfield{author}{\bibinfo{person}{Shenggui Li}, \bibinfo{person}{Fuzhao Xue},
  \bibinfo{person}{Chaitanya Baranwal}, \bibinfo{person}{Yongbin Li}, {and}
  \bibinfo{person}{Yang You}.} \bibinfo{year}{2021}\natexlab{}.
\newblock \showarticletitle{Sequence parallelism: Long sequence training from
  system perspective}.
\newblock \bibinfo{journal}{{\em arXiv preprint arXiv:2105.13120\/}}
  (\bibinfo{year}{2021}).
\newblock


\bibitem[\protect\citeauthoryear{Li, Zhao, Varma, Salpekar, Noordhuis, Li,
  Paszke, Smith, Vaughan, Damania, et~al\mbox{.}}{Li et~al\mbox{.}}{2020}]%
        {li2020pytorch}
\bibfield{author}{\bibinfo{person}{Shen Li}, \bibinfo{person}{Yanli Zhao},
  \bibinfo{person}{Rohan Varma}, \bibinfo{person}{Omkar Salpekar},
  \bibinfo{person}{Pieter Noordhuis}, \bibinfo{person}{Teng Li},
  \bibinfo{person}{Adam Paszke}, \bibinfo{person}{Jeff Smith},
  \bibinfo{person}{Brian Vaughan}, \bibinfo{person}{Pritam Damania},
  {et~al\mbox{.}}} \bibinfo{year}{2020}\natexlab{}.
\newblock \showarticletitle{Pytorch distributed: Experiences on accelerating
  data parallel training}.
\newblock \bibinfo{journal}{{\em arXiv preprint arXiv:2006.15704\/}}
  (\bibinfo{year}{2020}).
\newblock


\bibitem[\protect\citeauthoryear{Liang, Liu, Wright, Constable, Gu, Huang,
  Zhang, Feng, Huang, Wang, et~al\mbox{.}}{Liang et~al\mbox{.}}{2024}]%
        {liang2024torchtitan}
\bibfield{author}{\bibinfo{person}{Wanchao Liang}, \bibinfo{person}{Tianyu
  Liu}, \bibinfo{person}{Less Wright}, \bibinfo{person}{Will Constable},
  \bibinfo{person}{Andrew Gu}, \bibinfo{person}{Chien-Chin Huang},
  \bibinfo{person}{Iris Zhang}, \bibinfo{person}{Wei Feng},
  \bibinfo{person}{Howard Huang}, \bibinfo{person}{Junjie Wang},
  {et~al\mbox{.}}} \bibinfo{year}{2024}\natexlab{}.
\newblock \showarticletitle{TorchTitan: One-stop PyTorch native solution for
  production ready LLM pre-training}.
\newblock \bibinfo{journal}{{\em arXiv preprint arXiv:2410.06511\/}}
  (\bibinfo{year}{2024}).
\newblock


\bibitem[\protect\citeauthoryear{Liu, Feng, Wang, Wang, Liu, Zhao, Dengr, Ruan,
  Dai, Guo, et~al\mbox{.}}{Liu et~al\mbox{.}}{2024a}]%
        {liu2024deepseek}
\bibfield{author}{\bibinfo{person}{Aixin Liu}, \bibinfo{person}{Bei Feng},
  \bibinfo{person}{Bin Wang}, \bibinfo{person}{Bingxuan Wang},
  \bibinfo{person}{Bo Liu}, \bibinfo{person}{Chenggang Zhao},
  \bibinfo{person}{Chengqi Dengr}, \bibinfo{person}{Chong Ruan},
  \bibinfo{person}{Damai Dai}, \bibinfo{person}{Daya Guo}, {et~al\mbox{.}}}
  \bibinfo{year}{2024}\natexlab{a}.
\newblock \showarticletitle{Deepseek-v2: A strong, economical, and efficient
  mixture-of-experts language model}.
\newblock \bibinfo{journal}{{\em arXiv preprint arXiv:2405.04434\/}}
  (\bibinfo{year}{2024}).
\newblock


\bibitem[\protect\citeauthoryear{Liu, Feng, Xue, Wang, Wu, Lu, Zhao, Deng,
  Zhang, Ruan, et~al\mbox{.}}{Liu et~al\mbox{.}}{2024b}]%
        {liu2024deepseekv3}
\bibfield{author}{\bibinfo{person}{Aixin Liu}, \bibinfo{person}{Bei Feng},
  \bibinfo{person}{Bing Xue}, \bibinfo{person}{Bingxuan Wang},
  \bibinfo{person}{Bochao Wu}, \bibinfo{person}{Chengda Lu},
  \bibinfo{person}{Chenggang Zhao}, \bibinfo{person}{Chengqi Deng},
  \bibinfo{person}{Chenyu Zhang}, \bibinfo{person}{Chong Ruan},
  {et~al\mbox{.}}} \bibinfo{year}{2024}\natexlab{b}.
\newblock \showarticletitle{Deepseek-v3 technical report}.
\newblock \bibinfo{journal}{{\em arXiv preprint arXiv:2412.19437\/}}
  (\bibinfo{year}{2024}).
\newblock


\bibitem[\protect\citeauthoryear{Liu and Abbeel}{Liu and Abbeel}{2024}]%
        {liu2024blockwise}
\bibfield{author}{\bibinfo{person}{Hao Liu} {and} \bibinfo{person}{Pieter
  Abbeel}.} \bibinfo{year}{2024}\natexlab{}.
\newblock \showarticletitle{Blockwise Parallel Transformers for Large Context
  Models}.
\newblock \bibinfo{journal}{{\em Neural Information Processing Systems\/}}
  (\bibinfo{year}{2024}).
\newblock


\bibitem[\protect\citeauthoryear{Liu, Zaharia, and Abbeel}{Liu
  et~al\mbox{.}}{2023b}]%
        {liu2023ring}
\bibfield{author}{\bibinfo{person}{Hao Liu}, \bibinfo{person}{Matei Zaharia},
  {and} \bibinfo{person}{Pieter Abbeel}.} \bibinfo{year}{2023}\natexlab{b}.
\newblock \showarticletitle{Ring attention with blockwise transformers for
  near-infinite context}.
\newblock \bibinfo{journal}{{\em arXiv preprint arXiv:2310.01889\/}}
  (\bibinfo{year}{2023}).
\newblock


\bibitem[\protect\citeauthoryear{Liu, Wang, and Jiang}{Liu
  et~al\mbox{.}}{2023a}]%
        {liu2023janus}
\bibfield{author}{\bibinfo{person}{Juncai Liu}, \bibinfo{person}{Jessie~Hui
  Wang}, {and} \bibinfo{person}{Yimin Jiang}.}
  \bibinfo{year}{2023}\natexlab{a}.
\newblock \showarticletitle{Janus: A unified distributed training framework for
  sparse mixture-of-experts models}. In \bibinfo{booktitle}{{\em ACM SIGCOMM}}.
\newblock


\bibitem[\protect\citeauthoryear{Mahajan, Chu, Sridharan, and Akella}{Mahajan
  et~al\mbox{.}}{2023}]%
        {mahajan2023better}
\bibfield{author}{\bibinfo{person}{Kshiteej Mahajan},
  \bibinfo{person}{Ching-Hsiang Chu}, \bibinfo{person}{Srinivas Sridharan},
  {and} \bibinfo{person}{Aditya Akella}.} \bibinfo{year}{2023}\natexlab{}.
\newblock \showarticletitle{Better Together: Jointly Optimizing ML Collective
  Scheduling and Execution Planning using $\{$SYNDICATE$\}$}. In
  \bibinfo{booktitle}{{\em USENIX NSDI}}.
\newblock


\bibitem[\protect\citeauthoryear{Megatron-LM}{Megatron-LM}{2025}]%
        {megatron-lm}
Megatron-LM \bibinfo{year}{2025}\natexlab{}.
\newblock \bibinfo{title}{{GPU optimized techniques for training transformer
  models at-scale}}.
\newblock   (\bibinfo{year}{2025}).
\newblock
\newblock
\shownote{\url{https://github.com/NVIDIA/Megatron-LM}.}


\bibitem[\protect\citeauthoryear{Narayanan, Harlap, Phanishayee, Seshadri,
  Devanur, Ganger, Gibbons, and Zaharia}{Narayanan et~al\mbox{.}}{2019}]%
        {narayanan2019pipedream}
\bibfield{author}{\bibinfo{person}{Deepak Narayanan}, \bibinfo{person}{Aaron
  Harlap}, \bibinfo{person}{Amar Phanishayee}, \bibinfo{person}{Vivek
  Seshadri}, \bibinfo{person}{Nikhil~R Devanur}, \bibinfo{person}{Gregory~R
  Ganger}, \bibinfo{person}{Phillip~B Gibbons}, {and} \bibinfo{person}{Matei
  Zaharia}.} \bibinfo{year}{2019}\natexlab{}.
\newblock \showarticletitle{PipeDream: generalized pipeline parallelism for DNN
  training}. In \bibinfo{booktitle}{{\em ACM SOSP}}.
\newblock


\bibitem[\protect\citeauthoryear{Narayanan, Shoeybi, Casper, LeGresley,
  Patwary, Korthikanti, Vainbrand, Kashinkunti, Bernauer, Catanzaro,
  et~al\mbox{.}}{Narayanan et~al\mbox{.}}{2021}]%
        {narayanan2021efficient}
\bibfield{author}{\bibinfo{person}{Deepak Narayanan}, \bibinfo{person}{Mohammad
  Shoeybi}, \bibinfo{person}{Jared Casper}, \bibinfo{person}{Patrick
  LeGresley}, \bibinfo{person}{Mostofa Patwary}, \bibinfo{person}{Vijay
  Korthikanti}, \bibinfo{person}{Dmitri Vainbrand}, \bibinfo{person}{Prethvi
  Kashinkunti}, \bibinfo{person}{Julie Bernauer}, \bibinfo{person}{Bryan
  Catanzaro}, {et~al\mbox{.}}} \bibinfo{year}{2021}\natexlab{}.
\newblock \showarticletitle{Efficient large-scale language model training on
  gpu clusters using megatron-lm}. In \bibinfo{booktitle}{{\em International
  Conference for High Performance Computing, Networking, Storage and
  Analysis}}.
\newblock


\bibitem[\protect\citeauthoryear{NCCL}{NCCL}{2025}]%
        {nccl}
NCCL \bibinfo{year}{2025}\natexlab{}.
\newblock \bibinfo{title}{{Optimized primitives for inter-GPU communication}}.
\newblock   (\bibinfo{year}{2025}).
\newblock
\newblock
\shownote{\url{https://github.com/NVIDIA/nccl}.}


\bibitem[\protect\citeauthoryear{Nie, Zhao, Miao, Zhao, and Cui}{Nie
  et~al\mbox{.}}{2022}]%
        {nie2022hetumoe}
\bibfield{author}{\bibinfo{person}{Xiaonan Nie}, \bibinfo{person}{Pinxue Zhao},
  \bibinfo{person}{Xupeng Miao}, \bibinfo{person}{Tong Zhao}, {and}
  \bibinfo{person}{Bin Cui}.} \bibinfo{year}{2022}\natexlab{}.
\newblock \showarticletitle{HetuMoE: An efficient trillion-scale
  mixture-of-expert distributed training system}.
\newblock \bibinfo{journal}{{\em arXiv preprint arXiv:2203.14685\/}}
  (\bibinfo{year}{2022}).
\newblock


\bibitem[\protect\citeauthoryear{Pati, Aga, Islam, Jayasena, and Sinclair}{Pati
  et~al\mbox{.}}{2024}]%
        {pati2024t3}
\bibfield{author}{\bibinfo{person}{Suchita Pati}, \bibinfo{person}{Shaizeen
  Aga}, \bibinfo{person}{Mahzabeen Islam}, \bibinfo{person}{Nuwan Jayasena},
  {and} \bibinfo{person}{Matthew~D. Sinclair}.}
  \bibinfo{year}{2024}\natexlab{}.
\newblock \showarticletitle{T3: Transparent Tracking \& Triggering for
  Fine-grained Overlap of Compute \& Collectives}. In \bibinfo{booktitle}{{\em
  ACM ASPLOS}}.
\newblock


\bibitem[\protect\citeauthoryear{Peng, Wu, Wei, Zhao, Yang, Liu, Xiong, Yang,
  Ni, Hu, et~al\mbox{.}}{Peng et~al\mbox{.}}{2023}]%
        {peng2023fp8}
\bibfield{author}{\bibinfo{person}{Houwen Peng}, \bibinfo{person}{Kan Wu},
  \bibinfo{person}{Yixuan Wei}, \bibinfo{person}{Guoshuai Zhao},
  \bibinfo{person}{Yuxiang Yang}, \bibinfo{person}{Ze Liu},
  \bibinfo{person}{Yifan Xiong}, \bibinfo{person}{Ziyue Yang},
  \bibinfo{person}{Bolin Ni}, \bibinfo{person}{Jingcheng Hu}, {et~al\mbox{.}}}
  \bibinfo{year}{2023}\natexlab{}.
\newblock \showarticletitle{Fp8-lm: Training fp8 large language models}.
\newblock \bibinfo{journal}{{\em arXiv preprint arXiv:2310.18313\/}}
  (\bibinfo{year}{2023}).
\newblock


\bibitem[\protect\citeauthoryear{Peng, Zhu, Chen, Bao, Yi, Lan, Wu, and
  Guo}{Peng et~al\mbox{.}}{2019}]%
        {peng2019generic}
\bibfield{author}{\bibinfo{person}{Yanghua Peng}, \bibinfo{person}{Yibo Zhu},
  \bibinfo{person}{Yangrui Chen}, \bibinfo{person}{Yixin Bao},
  \bibinfo{person}{Bairen Yi}, \bibinfo{person}{Chang Lan},
  \bibinfo{person}{Chuan Wu}, {and} \bibinfo{person}{Chuanxiong Guo}.}
  \bibinfo{year}{2019}\natexlab{}.
\newblock \showarticletitle{A generic communication scheduler for distributed
  DNN training acceleration}. In \bibinfo{booktitle}{{\em ACM SOSP}}.
\newblock


\bibitem[\protect\citeauthoryear{Rajbhandari, Li, Yao, Zhang, Aminabadi, Awan,
  Rasley, and He}{Rajbhandari et~al\mbox{.}}{2022}]%
        {rajbhandari2022deepspeed}
\bibfield{author}{\bibinfo{person}{Samyam Rajbhandari},
  \bibinfo{person}{Conglong Li}, \bibinfo{person}{Zhewei Yao},
  \bibinfo{person}{Minjia Zhang}, \bibinfo{person}{Reza~Yazdani Aminabadi},
  \bibinfo{person}{Ammar~Ahmad Awan}, \bibinfo{person}{Jeff Rasley}, {and}
  \bibinfo{person}{Yuxiong He}.} \bibinfo{year}{2022}\natexlab{}.
\newblock \showarticletitle{{D}eep{S}peed-{M}o{E}: Advancing Mixture-of-Experts
  Inference and Training to Power Next-Generation {AI} Scale}. In
  \bibinfo{booktitle}{{\em International Conference on Machine Learning
  (ICML)}}.
\newblock


\bibitem[\protect\citeauthoryear{Rajbhandari, Rasley, Ruwase, and
  He}{Rajbhandari et~al\mbox{.}}{2020}]%
        {rajbhandari2020zero}
\bibfield{author}{\bibinfo{person}{Samyam Rajbhandari}, \bibinfo{person}{Jeff
  Rasley}, \bibinfo{person}{Olatunji Ruwase}, {and} \bibinfo{person}{Yuxiong
  He}.} \bibinfo{year}{2020}\natexlab{}.
\newblock \showarticletitle{Zero: Memory optimizations toward training trillion
  parameter models}. In \bibinfo{booktitle}{{\em International Conference for
  High Performance Computing, Networking, Storage and Analysis}}.
\newblock


\bibitem[\protect\citeauthoryear{Rajbhandari, Ruwase, Rasley, Smith, and
  He}{Rajbhandari et~al\mbox{.}}{2021}]%
        {rajbhandari2021zero}
\bibfield{author}{\bibinfo{person}{Samyam Rajbhandari},
  \bibinfo{person}{Olatunji Ruwase}, \bibinfo{person}{Jeff Rasley},
  \bibinfo{person}{Shaden Smith}, {and} \bibinfo{person}{Yuxiong He}.}
  \bibinfo{year}{2021}\natexlab{}.
\newblock \showarticletitle{Zero-infinity: Breaking the gpu memory wall for
  extreme scale deep learning}. In \bibinfo{booktitle}{{\em International
  Conference for High Performance Computing, Networking, Storage and
  Analysis}}.
\newblock


\bibitem[\protect\citeauthoryear{Rasley, Rajbhandari, Ruwase, and He}{Rasley
  et~al\mbox{.}}{2020}]%
        {rasley2020deepspeed}
\bibfield{author}{\bibinfo{person}{Jeff Rasley}, \bibinfo{person}{Samyam
  Rajbhandari}, \bibinfo{person}{Olatunji Ruwase}, {and}
  \bibinfo{person}{Yuxiong He}.} \bibinfo{year}{2020}\natexlab{}.
\newblock \showarticletitle{Deepspeed: System optimizations enable training
  deep learning models with over 100 billion parameters}. In
  \bibinfo{booktitle}{{\em ACM SIGKDD}}.
\newblock


\bibitem[\protect\citeauthoryear{Ren, Rajbhandari, Aminabadi, Ruwase, Yang,
  Zhang, Li, and He}{Ren et~al\mbox{.}}{2021}]%
        {ren2021zero}
\bibfield{author}{\bibinfo{person}{Jie Ren}, \bibinfo{person}{Samyam
  Rajbhandari}, \bibinfo{person}{Reza~Yazdani Aminabadi},
  \bibinfo{person}{Olatunji Ruwase}, \bibinfo{person}{Shuangyan Yang},
  \bibinfo{person}{Minjia Zhang}, \bibinfo{person}{Dong Li}, {and}
  \bibinfo{person}{Yuxiong He}.} \bibinfo{year}{2021}\natexlab{}.
\newblock \showarticletitle{Zero-offload: Democratizing billion-scale model
  training}. In \bibinfo{booktitle}{{\em USENIX ATC}}.
\newblock


\bibitem[\protect\citeauthoryear{Shazeer}{Shazeer}{2020}]%
        {shazeer2020glu}
\bibfield{author}{\bibinfo{person}{Noam Shazeer}.}
  \bibinfo{year}{2020}\natexlab{}.
\newblock \showarticletitle{Glu variants improve transformer}.
\newblock \bibinfo{journal}{{\em arXiv preprint arXiv:2002.05202\/}}
  (\bibinfo{year}{2020}).
\newblock


\bibitem[\protect\citeauthoryear{Shazeer, Mirhoseini, Maziarz, Davis, Le,
  Hinton, and Dean}{Shazeer et~al\mbox{.}}{2017}]%
        {shazeer2017outrageously}
\bibfield{author}{\bibinfo{person}{Noam Shazeer}, \bibinfo{person}{Azalia
  Mirhoseini}, \bibinfo{person}{Krzysztof Maziarz}, \bibinfo{person}{Andy
  Davis}, \bibinfo{person}{Quoc Le}, \bibinfo{person}{Geoffrey Hinton}, {and}
  \bibinfo{person}{Jeff Dean}.} \bibinfo{year}{2017}\natexlab{}.
\newblock \showarticletitle{Outrageously large neural networks: The
  sparsely-gated mixture-of-experts layer}.
\newblock \bibinfo{journal}{{\em arXiv preprint arXiv:1701.06538\/}}
  (\bibinfo{year}{2017}).
\newblock


\bibitem[\protect\citeauthoryear{Shen, Wu, Gong, Hao, Bai, Wu, Wu, Bian, Xiong,
  Yu, et~al\mbox{.}}{Shen et~al\mbox{.}}{2022}]%
        {shen2022se}
\bibfield{author}{\bibinfo{person}{Liang Shen}, \bibinfo{person}{Zhihua Wu},
  \bibinfo{person}{WeiBao Gong}, \bibinfo{person}{Hongxiang Hao},
  \bibinfo{person}{Yangfan Bai}, \bibinfo{person}{HuaChao Wu},
  \bibinfo{person}{Xinxuan Wu}, \bibinfo{person}{Jiang Bian},
  \bibinfo{person}{Haoyi Xiong}, \bibinfo{person}{Dianhai Yu}, {et~al\mbox{.}}}
  \bibinfo{year}{2022}\natexlab{}.
\newblock \showarticletitle{Se-moe: A scalable and efficient mixture-of-experts
  distributed training and inference system}.
\newblock \bibinfo{journal}{{\em arXiv preprint arXiv:2205.10034\/}}
  (\bibinfo{year}{2022}).
\newblock


\bibitem[\protect\citeauthoryear{Shoeybi, Patwary, Puri, LeGresley, Casper, and
  Catanzaro}{Shoeybi et~al\mbox{.}}{2019}]%
        {shoeybi2019megatron}
\bibfield{author}{\bibinfo{person}{Mohammad Shoeybi}, \bibinfo{person}{Mostofa
  Patwary}, \bibinfo{person}{Raul Puri}, \bibinfo{person}{Patrick LeGresley},
  \bibinfo{person}{Jared Casper}, {and} \bibinfo{person}{Bryan Catanzaro}.}
  \bibinfo{year}{2019}\natexlab{}.
\newblock \showarticletitle{Megatron-lm: Training multi-billion parameter
  language models using model parallelism}.
\newblock \bibinfo{journal}{{\em arXiv preprint arXiv:1909.08053\/}}
  (\bibinfo{year}{2019}).
\newblock


\bibitem[\protect\citeauthoryear{Touvron, Martin, Stone, Albert, Almahairi,
  Babaei, Bashlykov, Batra, Bhargava, Bhosale, et~al\mbox{.}}{Touvron
  et~al\mbox{.}}{2023}]%
        {touvron2023llama}
\bibfield{author}{\bibinfo{person}{Hugo Touvron}, \bibinfo{person}{Louis
  Martin}, \bibinfo{person}{Kevin Stone}, \bibinfo{person}{Peter Albert},
  \bibinfo{person}{Amjad Almahairi}, \bibinfo{person}{Yasmine Babaei},
  \bibinfo{person}{Nikolay Bashlykov}, \bibinfo{person}{Soumya Batra},
  \bibinfo{person}{Prajjwal Bhargava}, \bibinfo{person}{Shruti Bhosale},
  {et~al\mbox{.}}} \bibinfo{year}{2023}\natexlab{}.
\newblock \showarticletitle{Llama 2: Open foundation and fine-tuned chat
  models}.
\newblock \bibinfo{journal}{{\em arXiv preprint arXiv:2307.09288\/}}
  (\bibinfo{year}{2023}).
\newblock


\bibitem[\protect\citeauthoryear{TransformerEngine}{TransformerEngine}{2025}]%
        {transformer-engine}
TransformerEngine \bibinfo{year}{2025}\natexlab{}.
\newblock \bibinfo{title}{{A library for accelerating Transformer models on
  NVIDIA GPUs, including using 8-bit floating point (FP8) precision on Hopper
  and Ada GPUs, to provide better performance with lower memory utilization in
  both training and inference.}}
\newblock   (\bibinfo{year}{2025}).
\newblock
\newblock
\shownote{\url{https://github.com/NVIDIA/TransformerEngine}.}


\bibitem[\protect\citeauthoryear{Vaswani, Shazeer, Parmar, Uszkoreit, Jones,
  Gomez, Kaiser, and Polosukhin}{Vaswani et~al\mbox{.}}{2017}]%
        {vaswani2017attention}
\bibfield{author}{\bibinfo{person}{Ashish Vaswani}, \bibinfo{person}{Noam
  Shazeer}, \bibinfo{person}{Niki Parmar}, \bibinfo{person}{Jakob Uszkoreit},
  \bibinfo{person}{Llion Jones}, \bibinfo{person}{Aidan~N Gomez},
  \bibinfo{person}{{\L}ukasz Kaiser}, {and} \bibinfo{person}{Illia
  Polosukhin}.} \bibinfo{year}{2017}\natexlab{}.
\newblock \showarticletitle{Attention is all you need}.
\newblock \bibinfo{journal}{{\em Neural Information Processing Systems\/}}
  (\bibinfo{year}{2017}).
\newblock


\bibitem[\protect\citeauthoryear{Wang, Wei, Sabne, Davis, Ilbeyi, Hechtman,
  Chen, Murthy, Maggioni, Zhang, et~al\mbox{.}}{Wang et~al\mbox{.}}{2022}]%
        {wang2022overlap}
\bibfield{author}{\bibinfo{person}{Shibo Wang}, \bibinfo{person}{Jinliang Wei},
  \bibinfo{person}{Amit Sabne}, \bibinfo{person}{Andy Davis},
  \bibinfo{person}{Berkin Ilbeyi}, \bibinfo{person}{Blake Hechtman},
  \bibinfo{person}{Dehao Chen}, \bibinfo{person}{Karthik~Srinivasa Murthy},
  \bibinfo{person}{Marcello Maggioni}, \bibinfo{person}{Qiao Zhang},
  {et~al\mbox{.}}} \bibinfo{year}{2022}\natexlab{}.
\newblock \showarticletitle{Overlap communication with dependent computation
  via decomposition in large deep learning models}. In \bibinfo{booktitle}{{\em
  ACM ASPLOS}}.
\newblock


\bibitem[\protect\citeauthoryear{Zhang, Zheng, Lin, Jiang, Bao, Jiang, Hou,
  Cui, Zheng, Chang, et~al\mbox{.}}{Zhang et~al\mbox{.}}{2025a}]%
        {zhang2025comet}
\bibfield{author}{\bibinfo{person}{Shulai Zhang}, \bibinfo{person}{Ningxin
  Zheng}, \bibinfo{person}{Haibin Lin}, \bibinfo{person}{Ziheng Jiang},
  \bibinfo{person}{Wenlei Bao}, \bibinfo{person}{Chengquan Jiang},
  \bibinfo{person}{Qi Hou}, \bibinfo{person}{Weihao Cui}, \bibinfo{person}{Size
  Zheng}, \bibinfo{person}{Li-Wen Chang}, {et~al\mbox{.}}}
  \bibinfo{year}{2025}\natexlab{a}.
\newblock \showarticletitle{Comet: Fine-grained Computation-communication
  Overlapping for Mixture-of-Experts}.
\newblock \bibinfo{journal}{{\em arXiv preprint arXiv:2502.19811\/}}
  (\bibinfo{year}{2025}).
\newblock


\bibitem[\protect\citeauthoryear{Zhang, Zhong, Jiang, Hu, Sun, Ge, Zhu, Jiang,
  and Jin}{Zhang et~al\mbox{.}}{2025b}]%
        {zhang2025disttrain}
\bibfield{author}{\bibinfo{person}{Zili Zhang}, \bibinfo{person}{Yinmin Zhong},
  \bibinfo{person}{Yimin Jiang}, \bibinfo{person}{Hanpeng Hu},
  \bibinfo{person}{Jianjian Sun}, \bibinfo{person}{Zheng Ge},
  \bibinfo{person}{Yibo Zhu}, \bibinfo{person}{Daxin Jiang}, {and}
  \bibinfo{person}{Xin Jin}.} \bibinfo{year}{2025}\natexlab{b}.
\newblock \showarticletitle{Disttrain: Addressing model and data heterogeneity
  with disaggregated training for multimodal large language models}. In
  \bibinfo{booktitle}{{\em ACM SIGCOMM}}.
\newblock


\bibitem[\protect\citeauthoryear{Zhao, Gu, Varma, Luo, Huang, Xu, Wright,
  Shojanazeri, Ott, Shleifer, Desmaison, Balioglu, Damania, Nguyen, Chauhan,
  Hao, Mathews, and Li}{Zhao et~al\mbox{.}}{2023}]%
        {pytorch_fsdp}
\bibfield{author}{\bibinfo{person}{Yanli Zhao}, \bibinfo{person}{Andrew Gu},
  \bibinfo{person}{Rohan Varma}, \bibinfo{person}{Liang Luo},
  \bibinfo{person}{Chien-Chin Huang}, \bibinfo{person}{Min Xu},
  \bibinfo{person}{Less Wright}, \bibinfo{person}{Hamid Shojanazeri},
  \bibinfo{person}{Myle Ott}, \bibinfo{person}{Sam Shleifer},
  \bibinfo{person}{Alban Desmaison}, \bibinfo{person}{Can Balioglu},
  \bibinfo{person}{Pritam Damania}, \bibinfo{person}{Bernard Nguyen},
  \bibinfo{person}{Geeta Chauhan}, \bibinfo{person}{Yuchen Hao},
  \bibinfo{person}{Ajit Mathews}, {and} \bibinfo{person}{Shen Li}.}
  \bibinfo{year}{2023}\natexlab{}.
\newblock \showarticletitle{PyTorch FSDP: Experiences on Scaling Fully Sharded
  Data Parallel}.
\newblock \bibinfo{journal}{{\em Proceedings of the VLDB Endowment\/}}
  (\bibinfo{year}{2023}).
\newblock


\bibitem[\protect\citeauthoryear{Zheng, Fang, Zheng, Hou, Bao, Zheng, Jiang,
  Wang, Ye, Lin, et~al\mbox{.}}{Zheng et~al\mbox{.}}{2025}]%
        {zheng2025tilelink}
\bibfield{author}{\bibinfo{person}{Size Zheng}, \bibinfo{person}{Jin Fang},
  \bibinfo{person}{Xuegui Zheng}, \bibinfo{person}{Qi Hou},
  \bibinfo{person}{Wenlei Bao}, \bibinfo{person}{Ningxin Zheng},
  \bibinfo{person}{Ziheng Jiang}, \bibinfo{person}{Dongyang Wang},
  \bibinfo{person}{Jianxi Ye}, \bibinfo{person}{Haibin Lin}, {et~al\mbox{.}}}
  \bibinfo{year}{2025}\natexlab{}.
\newblock \showarticletitle{Tilelink: Generating efficient
  compute-communication overlapping kernels using tile-centric primitives}.
\newblock \bibinfo{journal}{{\em arXiv preprint arXiv:2503.20313\/}}
  (\bibinfo{year}{2025}).
\newblock


\end{thebibliography}

\clearpage

\balance

\appendix

% \parait{Appendices are supporting material that has not been peer-reviewed.}

\section{Appendix}
\label{apdx}

\subsection{Hierarchical Communication for Parameter Synchronization}
\label{apdx:hierarchical}

Let the full attention weights size be $P$, the dimension of model parallelism (TP or SP) be $n$, and the data parallel size be $d$. Typically, GPUs for model parallelism are located on the same node, requiring intra-node communication, whereas data parallelism spans across nodes, requiring inter-node communication. Consider a data parallelism group containing $d$ devices, each holding the identical partition of the parameter.

For parameter synchronization in TP attention, communication involves data of size $P/n$ across $d$ devices in two primary steps in LLM training:
\begin{itemize}[leftmargin=*]
    \item inter-node \texttt{reduce-scatter} operation, where the data size is $P/n$, on $d$ devices.
    \item inter-node \texttt{all-gather} operation, where the data size is $P/n$, on $d$ devices.
\end{itemize}
leading to primarily inter-node communication, with a communication volume of $2P/n(d-1)/d$.

With SP attention, the parameter synchronization involves the entire data of size $P$ across $n \times d$ devices. Considering the discrepancy between intra-node and inter-node network bandwidth, this process can be implemented by four-step hierarchical communication, where the replicated parameters are first reduced within a node and then reduced across nodes, before being distributed back to each device. Figure~\ref{fig:hierarchy}a illustrates a hierarchical communication example where $n=3$ and $d=2$. The detailed steps are as follows.
\begin{itemize}[leftmargin=*]
    \item intra-node \texttt{reduce-scatter} operation, where the data size is $P$, on $n$ devices.
    \item inter-node \texttt{reduce-scatter} operation, where the data size is $P/n$, on $d$ devices.
    \item inter-node \texttt{all-gather} operation, where the data size is $P/n$, on $d$ devices.
    \item intra-node \texttt{all-gather} operation, where the data size is $P$, on $n$ devices.
\end{itemize}
The inter-node communication volume in SP attention remains at $2P/n(d-1)/d$, with additional intra-node volume of $2P(n-1)/n$.  

Moreover, due to the distinct resources for intra-node and inter-node communications, these steps can be segmented into small chunks and pipelined to efficiently hide each other as shown in Figure~\ref{fig:hierarchy}b. The ratio of inter-node communication latency and intra-node communication latency is
\begin{align} \label{eq:ratio}
    \frac{1}{n} \times \frac{\text{intra-node bandwidth}}{\text{inter-node bandwidth}} \times \frac{n(d-1)}{d(n-1)}
\end{align}

Consider a typical training scenario involving an H100 SXM machine, where the NVLink bandwidth is 450 GB/s, and the inter-device NIC communication bandwidth is 50 GB/s. In this context, the latency of inter-node communication can easily surpass that of intra-node communication. This implies that the communication within a node can overshadow that between nodes. Consequently, in such scenarios, the synchronization of gradients and parameters with SP attention is, in fact, consistent with TP attention.

% \subsection{Selective Activation Rematerialization}
% \label{apdx:recomputation}
\end{sloppypar}
\end{document}